\documentclass{article}

\usepackage[square]{natbib}

\usepackage[preprint]{nips_2018}

\usepackage[utf8]{inputenc} % allow utf-8 input
\usepackage[T1]{fontenc}    % use 8-bit T1 fonts
\usepackage{hyperref}       % hyperlinks
\usepackage{url}            % simple URL typesetting
\usepackage{booktabs}       % professional-quality tables
\usepackage{amsfonts}       % blackboard math symbols
\usepackage{nicefrac}       % closed symbols for 1/2, etc.
\usepackage{microtype}      % microtypography

\usepackage{times}
\usepackage{epsfig}
\usepackage{graphicx}
\usepackage{amsmath}
\usepackage{amssymb}

\usepackage{amsthm}  
\usepackage{wasysym}

\usepackage{color}
\usepackage{dsfont}	
\usepackage{wrapfig}
\usepackage{footnote}
\usepackage{epstopdf}
\usepackage{enumitem}
\usepackage{subfigure}

\def\ie{\emph{i.e. }}

\makeatletter
\newcommand*{\rom}[1]{\expandafter\@slowromancap\romannumeral #1@}
\makeatother
\makeatletter
\newcommand\footnoteref[1]{\protected@xdef\@thefnmark{\ref{#1}}\@footnotemark}
\makeatother

\title{LMKL-Net: A Fast Localized Multiple Kernel Learning Solver via Deep Neural Networks}
\author{
	Ziming Zhang \\
	Mitsubishi Electric Research Laboratories (MERL) \\	
	Cambridge, MA 02139-1955 \\
	\texttt{zzhang@merl.com} \\
}

\begin{document}
	
\maketitle
	
\begin{abstract}

In this paper we propose solving localized multiple kernel learning (LMKL) using LMKL-Net, a feedforward deep neural network. In contrast to previous works, as a learning principle we propose {\em parameterizing} both the gating function for learning kernel combination weights and the multiclass classifier in LMKL using an attentional network (AN) and a multilayer perceptron (MLP), respectively. In this way we can learn the (nonlinear) decision function in LMKL (approximately) by sequential applications of AN and MLP. Empirically on benchmark datasets we demonstrate that overall LMKL-Net can not only outperform the state-of-the-art MKL solvers in terms of accuracy, but also be trained about {\em two orders of magnitude} faster with much smaller memory footprint for large-scale learning. %thanks to stochastic gradient descent (SGD). 

\end{abstract}
	
\section{Introduction}
Multiple kernel learning (MKL) is a classic yet powerful machine learning technique for integrating heterogeneous features in the reproducing kernel Hilbert space (RKHS) by linear or nonlinear combination of base kernels. It has been demonstrated successfully in many applications such as object detection in computer vision \cite{vedaldi2009multiple} as one of the joint winners in VOC2009 \cite{pascal-voc-2009}.

In the literature many MKL formulations have been proposed. For instance, \citet{bach2004multiple} proposed a block-$\ell_1$ regularized formulation for binary classification, and later \citet{kloft2011lp} proposed using $\ell_p$ norm as regularization in MKL. As a classic example we show the objective in SimpleMKL \cite{rakotomamonjy2008simplemkl} as follows:
\begin{align}\label{eqn:primal}
\min_{\mathbf{w}_m, \boldsymbol\beta\in\Delta^{M-1}, \boldsymbol\zeta, b} \frac{1}{2}\sum_{m=1}^M \frac{1}{\beta_m}\|\mathbf{w}_m\|_2^2 + C\sum_{i=1}^N \zeta_i, \; \mbox{s.t.} \; y_i\left(\sum_{m=1}^M\mathbf{w}_m^T\Phi_m(x_i) + b\right) \geq 1 - \zeta_i, \zeta_i\geq0, \forall i,
% & \hspace{4mm} \sum_{m=1}^M\beta_m=1, \beta_m\geq0, \forall m, \nonumber
\end{align}
where the set $\{(x_i, y_i)\}$ denote training data, $\mathbf{w}_m, \forall m$ and $b$ denote the classifier parameters, $\boldsymbol\beta=[\beta_m]\in\Delta^{M-1}$ denote the weights for $M$ kernels lying on the $(M-1)$-simplex space, $\boldsymbol\zeta=[\zeta_i]$ denote the slack variables for the hinge loss, $C\geq 0$ is a predefined regularization constant, $(\cdot)^T$ denotes the matrix transpose operator, and $\Phi_m(\cdot)$ denotes the feature map in RKHS for the $m$-th kernel $\mathbf{K}_m$ so that $\mathbf{K}_m(x_i, x_j) = \Phi_m(x_i)^T\Phi_m(x_i), \forall x_i, \forall x_j$. %To increase flexibility in controlling kernel weights, some other regularizers were proposed as well, such as $\ell_p$ norm \cite{kloft2011lp}. 

Based on the dual we can rewrite Eq. \ref{eqn:primal} as follows:
\begin{align}\label{eqn:wrapper}
& \min_{\boldsymbol\beta} J(\boldsymbol\beta) \; s.t. \; \sum_m\beta_m=1, \beta_m\geq 0, \forall m \\
& \mbox{where} \; J(\boldsymbol\beta) = \left\{ 
\begin{array}{l}\label{eqn:alpha}
\max_{\boldsymbol\alpha} -\frac{1}{2}\sum_{i,j}\alpha_i\alpha_jy_iy_j\sum_m\beta_m\mathbf{K}_m(x_i,x_j) + \sum_i\alpha_i \\ 
s.t. \; \sum_i\alpha_iy_i=0, 0\leq \alpha_i \leq C, \forall i. 
\end{array}
\right. 
\end{align}
Here $\alpha_i, \forall i$ denote the Lagrangian variables in the dual. To optimize Eq. \ref{eqn:wrapper} typically alternating optimization algorithms are developed, that is, learning $\boldsymbol\beta$ in Eq. \ref{eqn:wrapper} while fixing $\boldsymbol\alpha$ and learning $\boldsymbol\alpha$ in Eq. \ref{eqn:alpha} by solving a kernel support vector machine (SVM) problem while fixing $\boldsymbol\beta$. Such family of algorithms in MKL are called {\em wrapper methods}. From this perspective Eq. \ref{eqn:primal} essentially learns a kernel mixture by {\em linear} combination of base kernels. Accordingly we can write the {\em decision function}, $f$, in SimpleMKL for a new sample $z\in\mathcal{X}$ as follows:
\begin{align}\label{eqn:pred_score}
f_{Simple}(z) = \sum_i\alpha_iy_i\left[\sum_{m}\beta_m\mathbf{K}_m(z, x_i)\right].
\end{align}

% , for instance, and by taking the derivatives \wrt $\mathbf{w}_m$, we obtain 
% \begin{align}\label{eqn:w}
% \mathbf{w}_m=\sum_i \alpha_i y_i \eta_m(x_i) \Phi_m(x_i), \forall m,
% \end{align}
%  By substituding Eq. \ref{eqn:w} into Eq. \ref{eqn:LMKL}, we can rewrite the {\em prediction score}, $f(\tilde{x})$, for data $\tilde{x}$ as follows:
% \begin{align}\label{eqn:pred_score}
% \hspace{-2mm} f(\tilde{x}) = \sum_i\alpha_iy_i\left[\sum_{m=1}^M\eta_m(\tilde{x})\eta_m(x_i)\mathbf{K}_m(\tilde{x}, x_i)\right] + b.
% \end{align}

Localized MKL (LMKL) \cite{gonen08icml, lei2016localized, moeller2016unified} and its variants such as \cite{liu2014sample} are another family of MKL algorithms which learn {\em localized} (\ie data-dependent) kernel weights. For instance, \citet{gonen08icml} propose replacing $\boldsymbol\beta$ in Eq. \ref{eqn:primal} with a function $\eta$, leading to the following objective:
\begin{align}\label{eqn:LMKL}
\min_{\mathbf{w}_m, \eta_m(x)\geq0, \boldsymbol\zeta, b} \frac{1}{2}\sum_{m} \|\mathbf{w}_m\|_2^2 + C\sum_i \zeta_i, \;
\mbox{s.t.} \; y_i\left(\sum_{m}\eta_m(x_i)\mathbf{w}_m^T\Phi_m(x_i) + b\right) \geq 1 - \zeta_i, \zeta_i\geq0, \forall i, 
\end{align}
where $\eta_m(x_i)$ denotes the {\em gating function} that takes data $x_i$ as input and outputs the weight for the $m$-th kernel. The decision function is as follows:
\begin{align}\label{eqn:pred_score_LMKL}
f_{Loc}(z) = \sum_i\alpha_iy_i\left[\sum_{m}\eta_m(z)\mathbf{K}_m(z, x_i)\eta_m(x_i)\right] + b.
\end{align}

Comparing Eq. \ref{eqn:LMKL} with Eq. \ref{eqn:primal}, we can see that LMKL essentially {\em relaxes} the conventional MKL by introducing the data-dependent function $\eta$ in LMKL that may better capture data properties such as distributions for learning kernel weights. Since the learning capability of $\eta$ is high, previous works usually prefer regularizing it using explicit expressions such as Gaussian similarity in \cite{gonen08icml}.

Beyond learning linear mixtures of kernels, some works focus on learning nonlinear kernel combination such as involving product between kernels \cite{cortes2009learning, bach2009exploring, jawanpuria2015generalized, meirom2016nuc} or deep MKL \cite{strobl2013deep,  DBLP:journals/corr/abs-1709-10441} that embeds kernels into kernels. Among these works, the primal formulations may be nontrivial to write down explicitly. Instead some works preferred learning the decision functions directly based on Empirical Risk Minimization (ERM). For instance, in \cite{cortes2009learning, strobl2013deep} the nonlinear kernel mixtures are fed into the dual of an SVM to learn the parameters for both kernel mixtures and the classifiers. 
% This leads to the decision function, for instance in \cite{strobl2013deep}, as follows:
% \begin{align}\label{eqn:pred_score_DeepMKL}
% f_{Deep}(z) = \sum_i\alpha_iy_i\mathbf{K}_{\theta}(z, x_i) + b,
% \end{align}
% where $\theta$ denotes the parameter for the deep kernel mixtures over base kernels.
MKL, including multi-class MKL \cite{zien2007multiclass} and multi-label MKL \cite{ji2009multi,tang2009multiple}, can be further generalized to multi-task learning (MTMKL) \cite{jawanpuria2011multi, gonen2011multitask, murugesan2017multi}. However, these research topics are out of scope of this paper. %Different from MKL, in MTMKL each task has its own input data, each of which may be shared by different tasks. Kernels are computed using {\em task-specific} input data, but the kernel weights are shared by all the tasks. 

% Nowadays learning with large-scale data is attracting more and more attention. Such learning, however, brings new computational challenges to MKL. In terms of computational complexity, for instance, wrapper methods usually involve solving kernel SVMs with complexity of at least $O(N^2)$ where $N$ is the number of samples. In terms of memory footprint, the complexity is $O(MN^2)$ where $M$ is the number of kernels. As addressed in \cite{kloft2011lp} it takes about 3GB in memory to store one single dense real-number kernel matrix with 20K data samples, roughly speaking. Then with more samples or kernels, traditional algorithms will take much longer to return solutions or easily run out of memory. 
% %In addition, some state-of-the-art MKL solvers may have trouble in multi-class classification, such as SPG-GMKL \cite{Jain12} that consumes huge amount of memory. %For instance, to train SPG-MKL\footnote{We download the code from \url{http://www.cs.cornell.edu/~ashesh/pubs/code/SPG-GMKL/download.html}.} \cite{Jain12} using 10 precomputed kernels with small size 1530$\times$1530 and 102 classes\footnote{We download the kernels from \url{http://www.robots.ox.ac.uk/~vgg/software/MKL/}.}, we observed that it takes more than 256GB memory to run, which cannot be affordable on common PCs. 
% Therefore we believe that developing {\em efficient large-scale} MKL solvers will have significant algorithmic and practical impacts in the field. 

{\bf Motivation:} %To handle these challenges, we are exploring the solutions based on the following considerations:
From the dual perspective, solving kernel SVMs such as Eq. \ref{eqn:alpha} involves a constrained quadratic programming (QP) problem for computing the Lagrangian variables that determine {\em linear} decision boundaries in RKHS. This is the key procedure that consumes most of the computation. %From the perspective of functionality, QP learns  that are represented by the learned Lagrangian variables. To accelerate the learning we consider {\em preserving such functionality by learning a data-dependent function that can generate distinctive patterns/distributions for both positive and negative data}. In other words, rather than solving a QP problem, we would like to compute solutions directly from data as classifiers. 
Moreover, in the literature of LMKL it lacks of a principle for learning the gating functions rather than manually tuning the functions such as enforcing Gaussian distributions as prior. In addition the optimal decision functions may not be necessarily linear in RKHS for the sake of accuracy. Therefore, it is highly desirable to have principles for learning both gating and decision functions in LMKL.

Deep neural networks (DNNs) have been proven as a universal approximator for an arbitrary function \cite{mhaskar2016deep}. In fact recently researchers have started to apply DNNs as efficient solvers to some classic numerical problems such partial differential equations (PDEs) and backward stochastic differential equations (BSDEs) in high dimensional spaces \cite{weinan2017deep, sirignano2017dgm, long2017pde}. Conventionally solving these numerical problems with large amount of high dimensional data is very time-consuming. In contrast DNNs are able to efficiently output approximate solutions with sufficient precision. %In fact DNNs have been widely used in large-scale learning such as image classification \cite{krizhevsky2012imagenet}, thanks to stochastic gradient descent (SGD) and its variants as efficient training algorithms with linear complexity. The nature of stochastic approaches can leverage the memory usage in MKL as well.

%In this way, we essentially learn a bilinear localized prediction function (\eg Eq. \ref{eqn:pred_score}) for MKL whose variables are both data-dependent. Therefore, we propose a new challenging learning problem in MKL, namely bilaterally localized MKL (BLMKL). This problem (see formal definition in Sec.~\ref{sec:blmkl}) can be considered as a relaxed version of LMKL

{\bf Contributions:} Based on the considerations above, we propose a simple neural network, namely LMKL-Net, as an efficient solver for LMKL. As a learning principle we parameterize the gating function as well as the classifier in LMKL using an attentional network (AN) and a multilayer perceptron (MLP), respectively, without enforcing any (strong) prior explicitly. We expect that by fitting the data, the network can approximate both underlying optimal functions properly. The localized weights learned from AN guarantee that the kernel mixtures are indeed valid kernels. 
% To do so, we represent each data as a 2D matrix with size of number of data samples times number of kernels %\ie a slice from the kernel cube, 
% and feed it into LMKL-Net. %In order to map a 2D matrix into a scalar, MKL such as Eq. \ref{eqn:primal} or Eq. \ref{eqn:LMKL} fuses kernel information using kernel weights and then classifies the data. Correspondingly in DNNs 
% AN learns localized weights to map each matrix into a vector, on top of which MLP approximates decision boundaries in RKHS as the classifier. %Implicitly the functionalities of kernel weights for fusion and  Lagrangian variables for classification are naturally embedded in MKL-Net. %By integrating these two components together we can develop our DNNs for solving MKL.
% % there should exist two branches in the networks: one learns kernel weights for combination, and the other does the prediction. For each branch, we can implement it as a multilayer perceptron (MLP).
Empirically we demonstrate the superiority of LMKL-Net over the state-of-the-art MKL solvers in terms of accuracy, memory, and running speed.

\section{Related Work}
{\bf Localized MKL (LMKL):}
%In general LMKL relaxes the conventional MKL such as Eq. \ref{eqn:primal} by introducing a so-called gating function so that the kernel weights are determined data-dependently rather than constants. 
The success of LMKL highly depends on the gating functions. Due to different design choices, previous works can be categorized into either data/sample related \cite{gonen08icml, han2012probability, liu2014sample} or group/cluster related \cite{yang2009group, mu2011non, lei2016localized}. For instance, in \cite{gonen08icml} the gating function is defined explicitly as a Gaussian similarity function, while in \cite{lei2016localized} the gating function is represented as the likelihoods of data in the clusters that are predefined by the likelihood function. Recently \citet{moeller2016unified} proposed viewing the gating function in LMKL as an explicit feature map that can generate an additional kernel on the data. In all of these works, the classifier parameters are shared among all the data samples.

In contrast to previous works, we propose using an attentional network to approximate the unknown optimal gating function. This parameterization can avoid the need of any prior on the function.
%BLMKL learns not only localized kernel weights but also localized classifier parameters that are specific to every training and test data. From this perspective, BLMKL is a further relaxation of LMKL with more flexibility in learning models. Consequently solving BLMKL will be more difficult. 

{\bf Large-Scale MKL:}
%We refer to large scale mainly as the size of kernels rather than the number of kernels, because the cost of learning kernel weights is much smaller, in general. 
Several MKL solvers have addressed the large-scale learning problem from the perspective of either computational complexity or memory footprint, such as SILP-MKL \cite{sonnenburg2006large}, SimpleMKL \cite{rakotomamonjy2008simplemkl}, $\ell_p$-norm-MKL \cite{kloft2011lp}, GMKL \cite{varma2009more}, SPG-GMKL \cite{Jain12}, UFO-MKL \cite{orabona2011ultra}, OBSCURE \cite{orabona2012multi}, and MWUMKL \cite{moeller2014geometric}. None of them, however, is proposed for LMKL.

Differently, LMKL-Net is able to solve the large-scale LMKL problem efficiently with much smaller memory footprint, thanks to stochastic gradient descent (SGD). Empirically we observe significant speedup using LMKL-Net, compared with the solvers above.
%based on dimension reduction to generate low-dimensional representations from large-scale kernels. Such procedure is explicitly embedded in the network. Furthermore SGD in training MKL-Net is good at leveraging computation as well as memory for large-scale learning. % In our experiments we will compare MKL-Net with these solvers to demonstrate our superiority.

{\bf Optimization:}
Wrapper methods are widely used in MKL that actually alternate the optimization between solving (multi-class) SVM problems and updating the kernel weights. Such methods include semi-infinite linear program (SILP) \cite{sonnenburg2006large}, reduced gradient \cite{rakotomamonjy2008simplemkl}, LPBoost \cite{gehler2009feature}, Newton's method \cite{kloft2011lp}, mirror descent \cite{jagarlapudi2009algorithmics}, spectral projected gradient (SPG) \cite{Jain12}, and triply stochastic gradients \cite{li2017triply}. \cite{gonen2011multiple, bucak2014multiple} provided nice reviews on different MKL algorithms. %Since solving kernel SVMs involves quadratic programming (QP), 
Such methods, however, cannot be scalable well in general without using some clever implementation techniques such as computing kernels on the fly \cite{Jain12,li2017triply}.

To overcome this problem, online learning based approaches have been proposed for MKL that have much lower memory requirement. For instance, \citet{martins2011online}, \citet{orabona2011ultra}, \citet{orabona2012multi} and \citet{ijcai2017-758} proposed utilizing stochastic gradient descent (SGD) to optimize the primal of MKL. \citet{alioschamultiple} proposed a multiple epochs of stochastic variance reduced gradient (SVRG) approach for $\ell_p$-norm MKL.

In our work we employ SGD to train LMKL-Net as well that can only have weak convergence in probability \cite{bottou2016optimization}, due to the non-convexity of our network. We notice that very recently \citet{song2017optimizing} proposed a deep kernel machine optimization (DKMO) framework that embeds kernel matrices using Nystr\"{o}m kernel approximations and learns task-specific representations through the fusion of multiple embeddings using deep learning. As a classifier it has been demonstrated that empirically DKMO can improve performance. It is very unclear, however, whether DKMO indeed solves an MKL problem in terms of optimization. On the contrary our LMKL-Net is a valid solver for LMKL.

\section{LMKL-Net}\label{sec:mkl-net}
\subsection{Mathematical Modeling}
{\bf Key Notations:}
We denote $\mathbf{K}(x)\in\mathbb{R}^{N\times M}, \forall x\in\mathcal{X}$ as a matrix for data sample $x$ with $N$ training samples and $M$ kernels, $y\in\mathcal{Y}$ as the class label of $x$, %$\mathbf{K}_m(x_i)\in\mathbb{R}^N, \forall m$ as the $i$-th column in the matrix for the $m$-th kernel, 
$h: \mathbb{R}^{N\times M}\times\Omega\rightarrow\Delta^{N\times M}$ as a function parameterized by $\Omega$ for learning localized kernel weights on a $(NM-1)$-simplex, $g: \mathbb{R}^N\times\Pi\rightarrow\mathbb{R}^{|\mathcal{Y}|}$ as a classifier parameterized by $\Pi$ for $|\mathcal{Y}|$ classes, and $\ell$ as a loss function. %Note that we assume that we do not have the access to the original data $x$.

{\bf Joint Gating Function:}
Letting $\mathbf{K}_{\eta_m}(z, x_i)=\eta_m(z)\eta_m(x_i)$ denote the kernel defined by the gating function $\eta_m$, we then can rewrite the kernel mixture in Eq. \ref{eqn:LMKL} as $\sum_m\eta_m(z)\mathbf{K}_m(z, x_i)\eta_m(x_i) = \sum_m\mathbf{K}_m(z, x_i)\mathbf{K}_{\eta_m}(z, x_i)$. This perspective of LMKL has been explored in \citet{moeller2016unified}. However, there are two major difficulties to compute the gating function $\eta_m$. First of all we may require the access to original data, which may not be available. Secondly in each update of function $\eta_m, \forall m\in[M]$ we have to recalculate $\eta_m(x_i), \forall x_i\in\mathcal{X}$, which may be very time consuming and memory inefficient. 

To overcome these problems in conventional LMKL we are inspired by the kernel mixture in Eq. \ref{eqn:pred_score} and propose a new gating function defined on not only $z$ but also $x_i$, namely $\eta_m(z, x_i)$, and define our kernel mixture as $\sum_m\eta_m(z,x_i)\mathbf{K}_m(z, x_i)\eta_m(z,x_i)=\sum_m\eta_m(z,x_i)^2\mathbf{K}_m(z, x_i)$. In contrast to Eq. \ref{eqn:pred_score} where the kernel weights in $\boldsymbol\beta$ are constant for all the data, now we would like to learn localized weights based on our joint gating function. In this way we can view $\eta_m(z,x_i)$ as a joint function over variables $z$ and $x_i$. This shares some similarities with the input kernel matrices that inspire us to learn (approximately) function $\eta_m, \forall m$ based on the kernels.

{\bf Decision Function:} To further generalize the classifier from linear to nonlinear for an arbitrary data sample $z$, we propose our decision function as follows:
\begin{align}\label{eqn:f}
f(z; \omega, \pi) = g\left(\sum_{m=1}^M h\Big(\mathbf{K}(z); \omega\Big)\otimes\mathbf{K}(z); \pi\right), 
\end{align}
where function $h$ is used to approximate the {\em normalized square} of our joint gating function, that is,
\begin{align}\label{eqn:h}
\left[h\Big(\mathbf{K}(z); \omega\Big)\right]_{i,m}\simeq\frac{\eta_m(z,x_i)^2}{\sum_{m,i}\eta_m(z,x_i)^2}, \forall m, \forall i,
\end{align}
for the entry at $(i,m)$ in the matrix, $\omega\in\Omega, \pi\in\Pi$ are the parameters of $h, g$, respectively, and $\otimes$ denotes the entry-wise product operator. %Note that our decision function may be related to Eq. \ref{eqn:pred_score_DeepMKL} for deep MKL as well, if the outputs of $h$ form a valid kernel matrix.

%{\bf Relation to Deep Kernel Learning:} Conventional deep kernel approaches learn {\em explicit} kernel mixture functions by, for instance, predefining the hierarchical architectures (\ie fixed rules). In contrast, BLMKL is capable to learn {\em implicit} kernel mixture functions by constructing $f_{BL}$ in Eq.~\ref{eqn:mkl} as a nonlinear classifier, \ie introducing more kernel operations such as multiplication between kernels into either $\pi$ or $\eta$ or both. %We can also construct $f$ based on fixed rules. From this perspective BLMKL may provide us an efficient way to learn deep kernels.

{\bf Problem Definition:}
In this paper we only focus on the MKL problems where only kernels and data labels are provided {\em with no access} to the original data. This setup is heavily used and studied in the literature \cite{gonen2011multiple}, and thus we only compare our approach with MKL algorithms (see our experimental section in Sec. \ref{sec:exp}) to emphasize the effectiveness of LMKL-Net in solving MKL problems.

Similar to \cite{cortes2009learning, strobl2013deep}, we would like to learn our decision function based on ERM as follows:
\begin{align}
\min_{\omega\in\Omega, \pi\in\Pi}\sum_{i}\ell\Big(y_i, f(x_i; \omega, \pi)\Big),
\end{align}
where $\Omega, \Pi$ denote the feasible parameter spaces that may be restricted by some constraints such as regularization. For simplification in expression we assume that all the parameter constraints have been embedded in the feasible spaces implicitly.

\begin{figure}[t]
    \begin{center}
        \centerline{\includegraphics[width=\columnwidth]{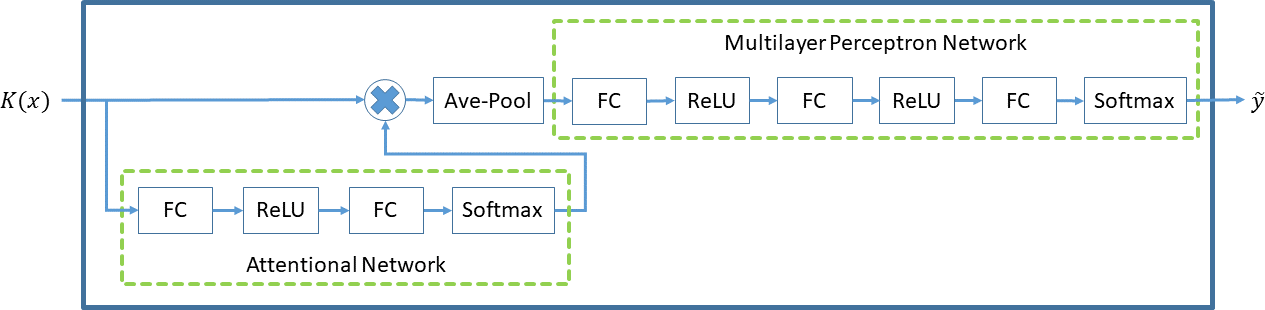}}
    \end{center}
	\vspace{-5mm}
	\caption{\footnotesize LMKL-Net architecture.}\label{fig:network}
    %\vspace{-3mm}
\end{figure}

\subsection{Network Architecture}

Now we would like to parameterize the functions $h,g$ in Eq. \ref{eqn:f} using DNNs. Fig. \ref{fig:network} illustrates the architecture of our LMKL-Net, where ``FC'' denotes a fully connected layer, ``ReLU'' denotes rectified linear units \cite{nair2010rectified} as the activation function, ``Ave-Pool'' denotes the average pooling operation over the kernels, and ``Softmax'' denotes a softmax function to normalize the inputs. Accordingly the parameter $\omega$ in $h$ is determined by the two FC layers in the attentional network (AN), and the parameter $\pi$ in $g$ is determined by the three FC layers in the multilayer perceptron network (MLP), respectively. By sequentially applying AN, average pooling, and MLP we indeed compute the decision function in Eq. \ref{eqn:f}. Empirically we find that ReLU leads to better performance with faster convergence than other activation functions, and adding more ``ReLU$\rightarrow$FC'' layers into the architecture, however, does not increase accuracy necessarily or significantly, but leads to more computational burden.

We train the network using ADAM \cite{kingma2014adam}, a variant of SGD with adaptive learning rates, with mini-batches to learn both kernel weights and classifier parameters {\em simultaneously}, in contrast to the conventional MKL algorithms such as wrapper methods. LMKL-Net can achieve {\em weak} convergence in probability based on the analysis in \cite{bottou2016optimization} for nonconvex optimization using SGD. %Empirically we observe that the nonlinearity in our solvers helps accelerate the convergence significantly as well as improve the accuracy.
Generally speaking, the computational complexity of LMKL-Net is {\em linear} to the number of its parameters, the size of mini-batches, and the number of iterations.

\begin{figure}[t]
    \begin{center}
        \centerline{\includegraphics[width=\columnwidth]{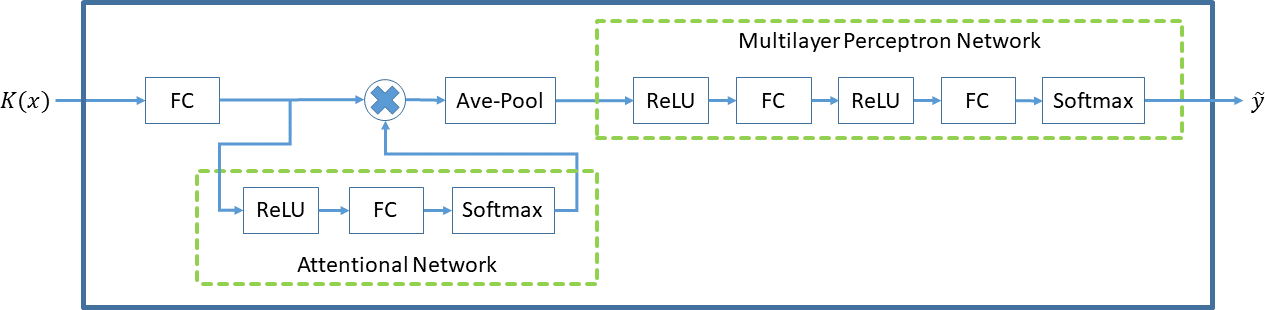}}
    \end{center}
	\vspace{-5mm}
	\caption{\footnotesize An equivalent architecture for accelerating training and inference of LMKL-Net by sharing weights.}\label{fig:network2}
    \vspace{-4mm}
\end{figure}

{\bf Training \& Inference Acceleration:}
In large-scale problems where the training samples are dominant, most of the computation for optimizing LMKL-Net in Fig. \ref{fig:network} are spent on the two FC layers in AN and the first FC layer in MLP. To accelerate both training and inference, we enforce that the first FC layers in both AN and MLP share the same weights. This transfers Fig. \ref{fig:network} to a new network architecture as illustrated in Fig. \ref{fig:network2}. Empirically we observe almost identical accuracy on different datasets using both architectures, while the new architecture is trained significantly faster. Therefore in our experiments we report our performance based on the architecture in Fig. \ref{fig:network2}.

{\bf Discussion:}
The idea of neural network based LMKL solver can be utilized for solving some other MKL learning problems. For instance, in order to solve SimpleMKL in Eq. \ref{eqn:pred_score} we can feed a constant vector into AN to learn $\boldsymbol\beta$ and use an FC layer with non-negative weights as a constraint instead of MLP to learn $\boldsymbol\alpha$. We observe that in terms of accuracy in practice this network is always worse than LMKL-Net. Sparse learning in kernel SVMs are useful in practice to reduce the size of kernel matrix per data. To mimic this nice property, we can add the group sparsity into the learning of the first FC layer in Fig. \ref{fig:network2}. However, the investigation on these specific network design choices is out of scope of this paper, and we will consider them in our future work.

\section{Experiments}\label{sec:exp}
%We test LMKL-Net on binary and multiclass classification problems given multiple kernels.

{\bf Implementation:}
In our experiments, we utilize the cross entropy loss because it performs superiorly than other losses and can handle multiclass classification problems inherently. We set the output dimension in the first FC layer in Fig. \ref{fig:network2} to 256D, and keep using it in MLP. The output dimension for the FC layer in AN is $N$ so that the entry-wise product operator can work. In our experiments using cross-validation we observe that higher dimensions than 256D do not necessarily lead to better accuracy but with remarkably longer running time, and the accuracy using lower dimensions starts to decrease. We initialize all FC layers randomly based on Gaussian distributions. We also observe that in our experiments having smaller weight decay \cite{krogh1992simple} has little impact on the accuracy while larger weight decay worsens the performance significantly. Therefore we decide not to include weight decay in our experiments. We train the network for 200 epochs with batch size 256 and learning rate 0.001 for ADAM optimizer in DNN training. All these numbers are determined by cross-validation. We also utilize batch-normalization to accelerate the training. The results are reported based on 3 runs.

{\bf Benchmarks:}
For binary problems we test on 6 datasets, \ie adult-8, news20, phishing, rcv1, real-sim, and w7a. For multiclass problems we test on 8 datasets, \ie aloi, covtype, letter, protein, sensit (combined), sensorless, shuttle, and SVHN. We download all the datasets (scaled versions if available) from \url{https://www.csie.ntu.edu.tw/~cjlin/libsvmtools/datasets/} which also lists the statistical information of the datasets such as numbers of training/validation/test data, classes and dimension of features. Please refer to the website for more dataset details. 

Considering the memory limit we intentionally control the number of data samples for generating kernels. Specifically in each dataset we randomly select 20K samples from training/validation data if the samples are more than 20K, otherwise we use all of them. Similarly for test data, we randomly select 10K samples if there are more, otherwise use all. Then by referring to VGG MKL dataset at \url{http://www.robots.ox.ac.uk/~vgg/software/MKL/} which consists of 10 Gaussian RBF kernels, we create 10 RBF kernels as well using the selected samples. We determine the window sizes in RBF-kernels to be proportional to the maximum Euclidean distance among the training features from 0.1 to 1 step by 0.1. We precompute all the kernels as the inputs for all the solvers.

To measure the classification performance, we utilize accuracy that is defined as the number of correctly classified samples divided by the total number of testing samples. The training and testing data samples have been balanced roughly.

{\bf Competitors:} 
We compare LMKL-Net with some state-of-the-art MKL solvers with public code that can directly take kernels as input. We tune each solver so that we can report the best performance that we can achieve.

For binary classification, we test GMKL \cite{varma2009more}\footnote{\url{http://www.cs.cornell.edu/~ashesh/pubs/code/SPG-GMKL/download.html}}, SMO-MKL \cite{Vishy10}\footnotemark[1], SPG-GMKL \cite{Jain12}\footnotemark[1], LMKL \cite{gonen08icml}\footnote{\url{http://users.ics.aalto.fi/gonen/icml08.php}}, Lp-MKL \cite{kloft2011lp}\footnote{\url{http://doc.ml.tu-berlin.de/nonsparse_mkl}}, UFO-MKL \cite{orabona2011ultra}\footnote{\url{https://github.com/denizyuret/dogma/tree/master/demos}}, OBSCURE \cite{orabona2010online}\footnotemark[4], and UNIFORM \cite{UNIFORM} (\ie average kernels with SVMs). We observe that GMKL, SMO-MKL, and SPG-GMKL always return identical results with different running speed, and thus report their results under the name of SPG-GMKL since it is fastest. Similarly for the two online learning methods, in our experiments OBSCURE outperforms UFO-MKL significantly in terms of both accuracy and running speed. Therefore, we only report the results of OBSCURE. We also observe that too much effort is needed such as heavily tuning parameters to make Lp-MKL work on our data and the results are often worse than others. %Therefore we will not report these results.
For multiclass classification, we compare LMKL-Net with OBSCURE and UNIFORM because we find that most of existing code that we use cannot handle multiclass classification properly. 

Note that among all the competitors OBSCURE is the most related to our LMKL-Net in terms of linear complexity in learning as well as its classification performance. OBSCURE utilizes group sparsity on classifier parameters  (\ie $\|\mathbf{w}\|_{2,p}, 1<p\leq2$) as regularization and proposes a two-stage online learning algorithm (first online then batch) to optimize the primal formulation with complicate calculation. In contrast, LMKL-Net is a network based solver that can be efficiently trained using SGD. This remarkable difference in optimization leads to the fact that empirically LMKL-Net can be trained significantly faster with much smaller memory demand.

\setlength{\tabcolsep}{4pt}
\begin{table*}[t]\small
	\begin{center}    
		\begin{tabular}{|c|c|c|c|c|c|c|c|}
			\hline & adult-8 & news20 & phishing & rcv1 & real-sim & w7a & average \\ 
            \hline UNIFORM & 81.94 & 93.33 & 46.16 & 96.37 & 96.56 & 90.37 & 84.12 \\
            \hline SPG-GMKL & 84.13 & 90.27 & 95.26 & 95.57 & 92.21 & 97.05 & 92.42 \\
            \hline LMKL & 78.09 & {\bf 95.52} & 52.04 & {\bf 96.76} & {\bf 97.09} & 97.75 & 86.21 \\
            \hline Lp-MKL & 76.33 & - & - & - & - & 97.05 & - \\
            \hline OBSCURE & 84.22$\pm$0.09 & 94.68$\pm$0.03 & 97.25$\pm$0.00 & 96.55$\pm$0.05 & 96.94$\pm$0.00 & 98.50$\pm$0.02 & 94.69\\
            \hline
%            \hline MKL-Net & {\bf 85.46$\pm$0.16} & 92.37$\pm$0.40 & {\bf 98.73$\pm$0.14} & {\bf 97.36$\pm$0.03} & 96.34$\pm$0.08 & {\bf 98.79$\pm$0.03} & {\bf 94.84} \\
            \hline {\bf Ours} & {\bf 84.62$\pm$0.15} & 93.53$\pm$0.03 & {\bf 98.17$\pm$0.25} & 96.71$\pm$0.07 & 96.42$\pm$0.09 & {\bf 98.74$\pm$0.03} & {\bf 94.70} \\
			\hline
		\end{tabular}
	\end{center}
    \vspace{-2mm}
	\caption{\footnotesize Comparison on binary classification accuracy ($\%$). ``-'' indicates that we cannot achieve reasonable performance on the dataset.}
	\label{tab:binary}
    \vspace{-3mm}
\end{table*}

\begin{figure*}[t]
    \begin{minipage}[b]{0.325\linewidth}
		\begin{center}
			\centerline{\includegraphics[width=\columnwidth]{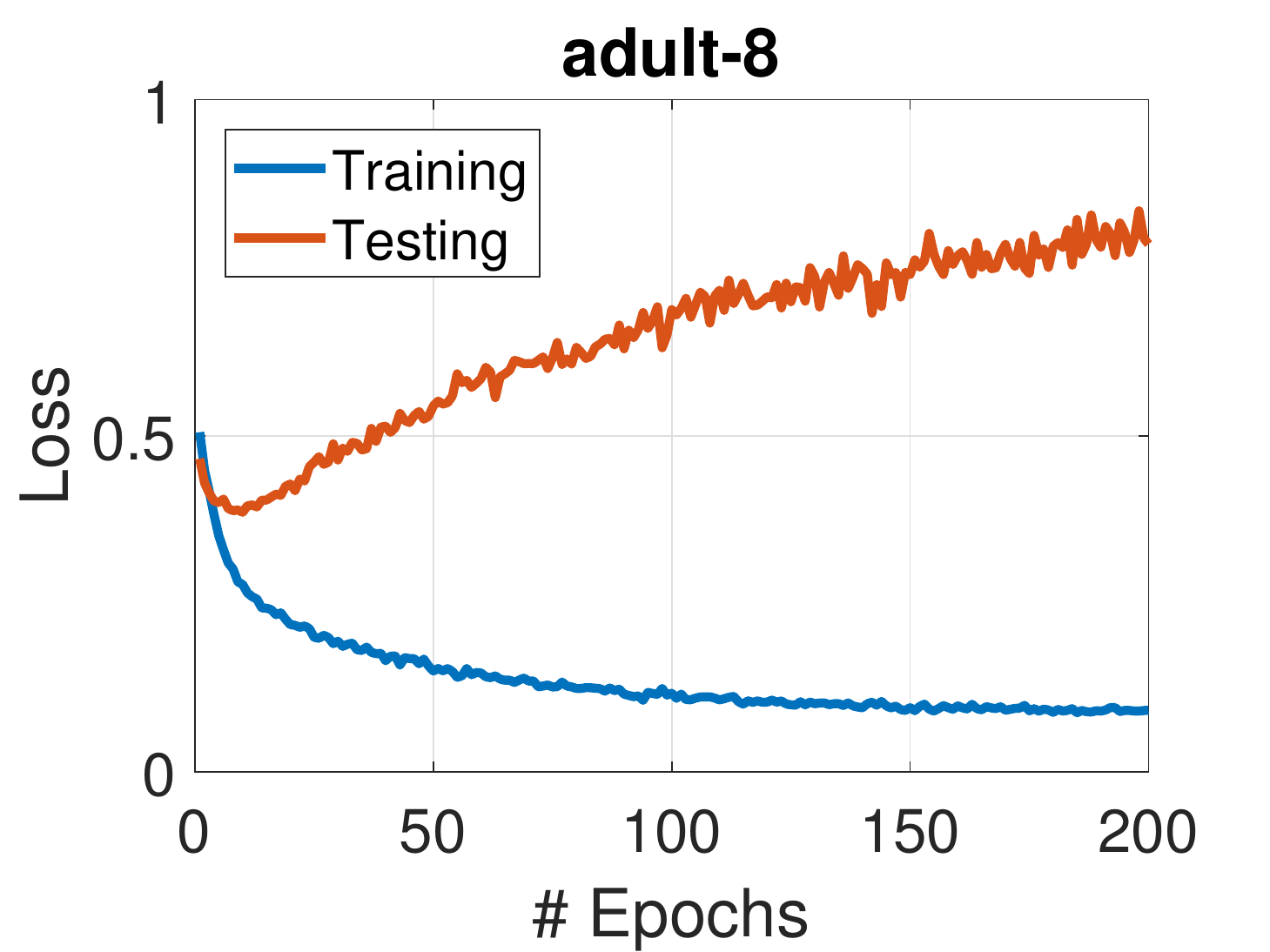}}
%			\centerline{\footnotesize{(a) Sampling using Eq. \ref{eqn:GD-sampler}}}
		\end{center}
	\end{minipage}
	\begin{minipage}[b]{0.325\linewidth}
		\begin{center}
			\centerline{\includegraphics[width=\columnwidth]{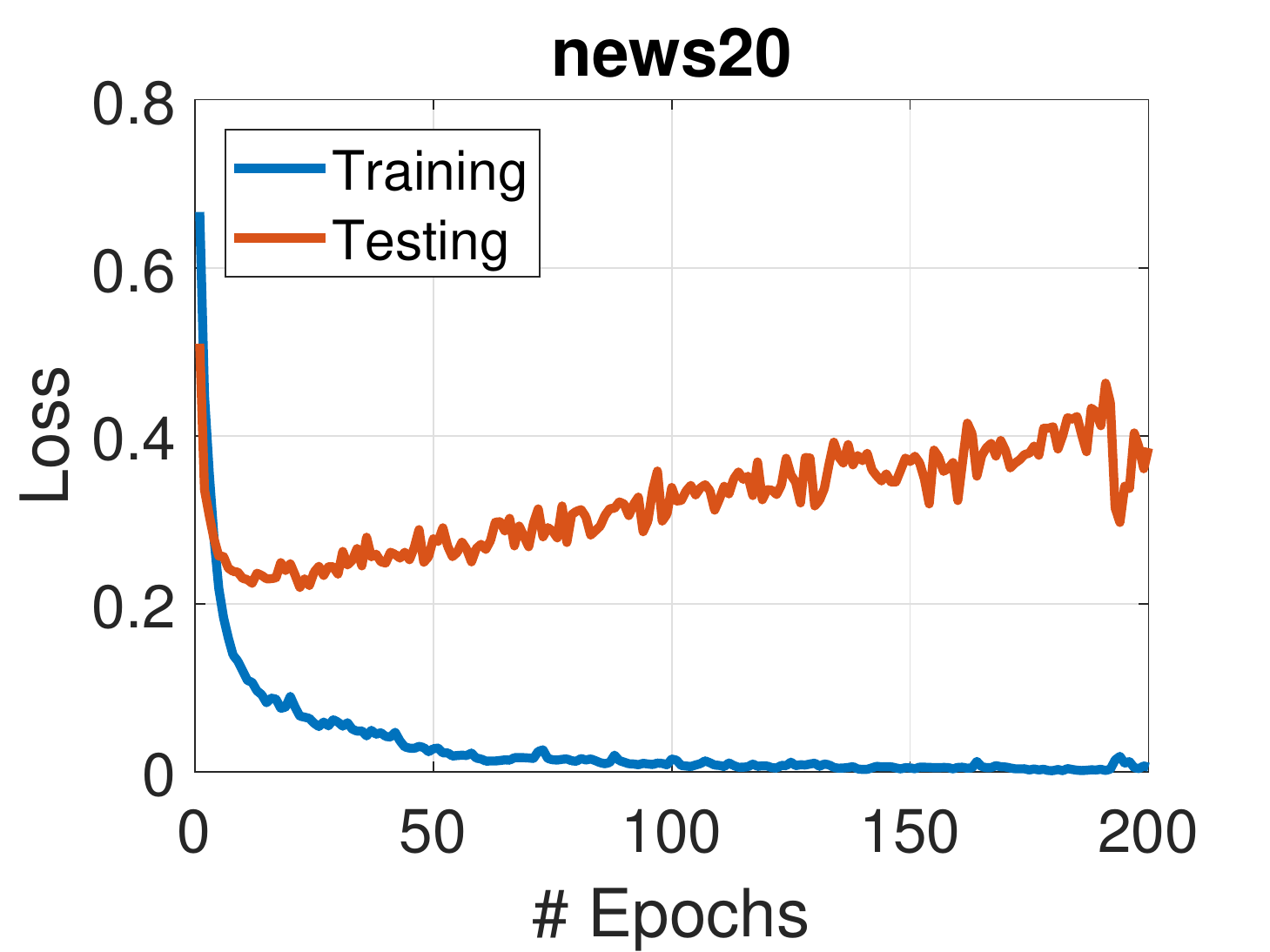}}
%			\centerline{\footnotesize{(b) Sampling using Eq. \ref{eqn:eta_t}}}
		\end{center}
	\end{minipage}	
	\begin{minipage}[b]{0.325\linewidth}
		\begin{center}
			\centerline{\includegraphics[width=\columnwidth]{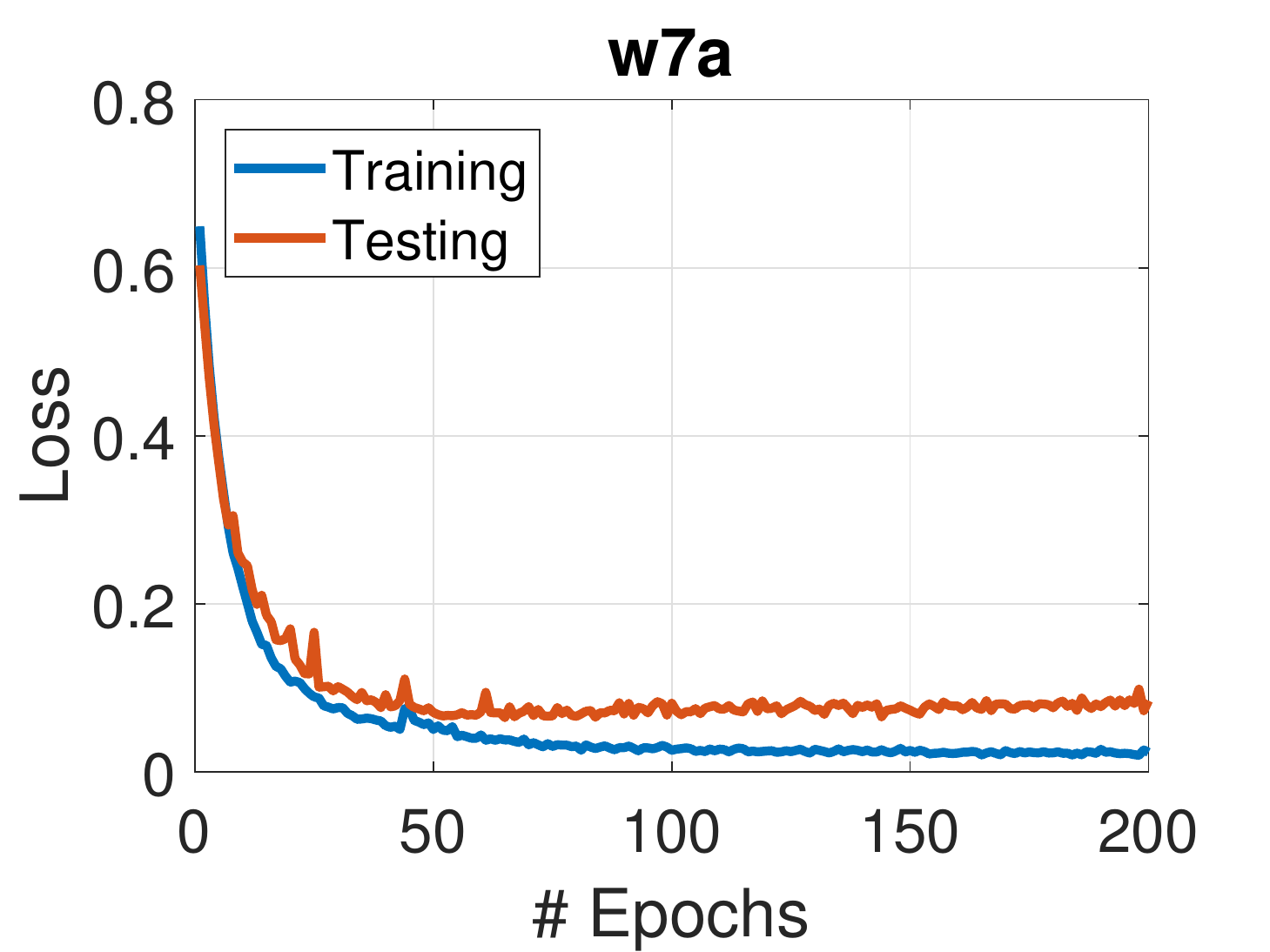}}
%			\centerline{\footnotesize{(b) Sampling using Eq. \ref{eqn:eta_t}}}
		\end{center}
	\end{minipage}	
    \begin{minipage}[b]{0.325\linewidth}
		\begin{center}
			\centerline{\includegraphics[width=\columnwidth]{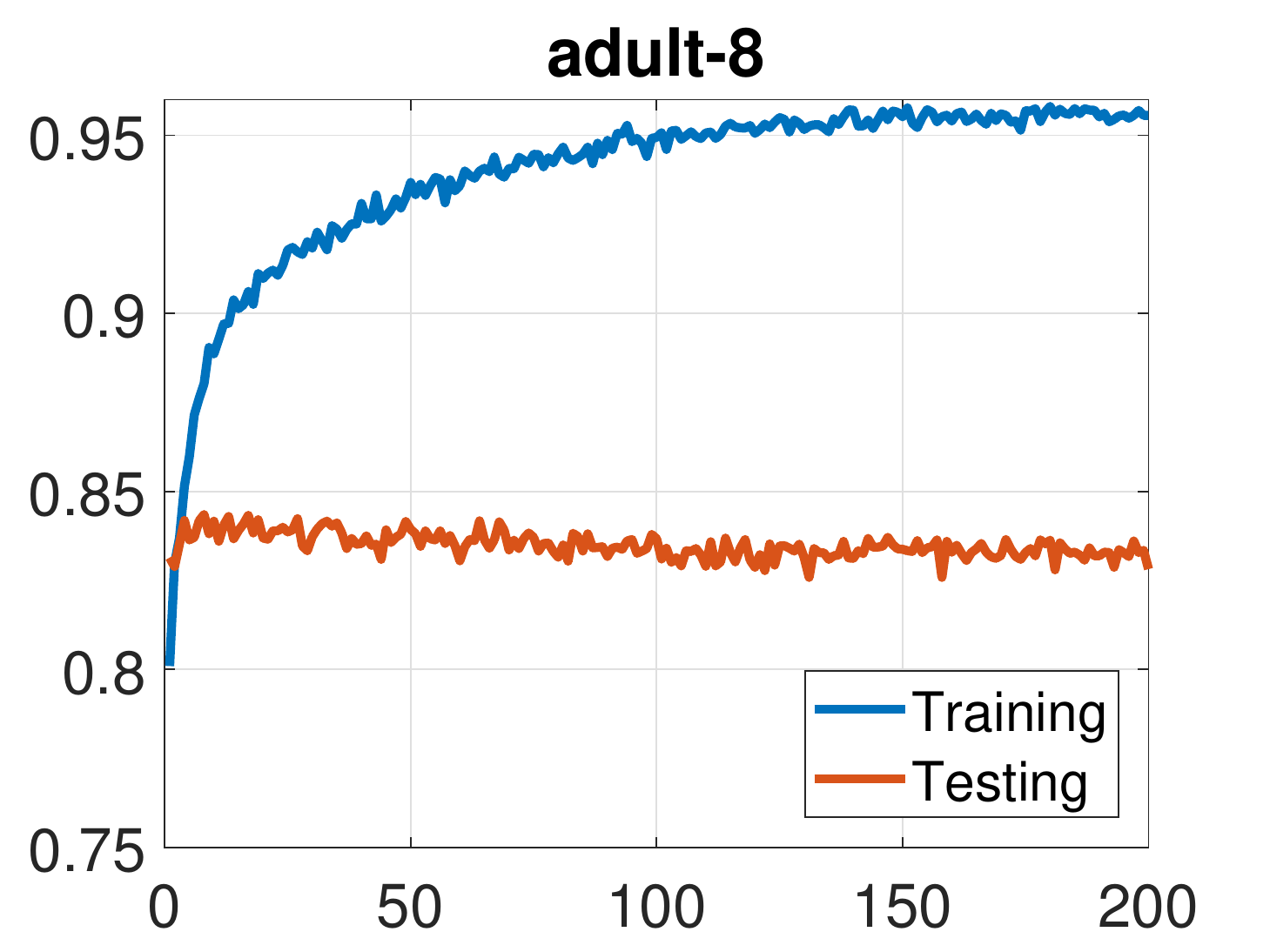}}
%			\centerline{\footnotesize{(a) Sampling using Eq. \ref{eqn:GD-sampler}}}
		\end{center}
	\end{minipage}
	\begin{minipage}[b]{0.325\linewidth}
		\begin{center}
			\centerline{\includegraphics[width=\columnwidth]{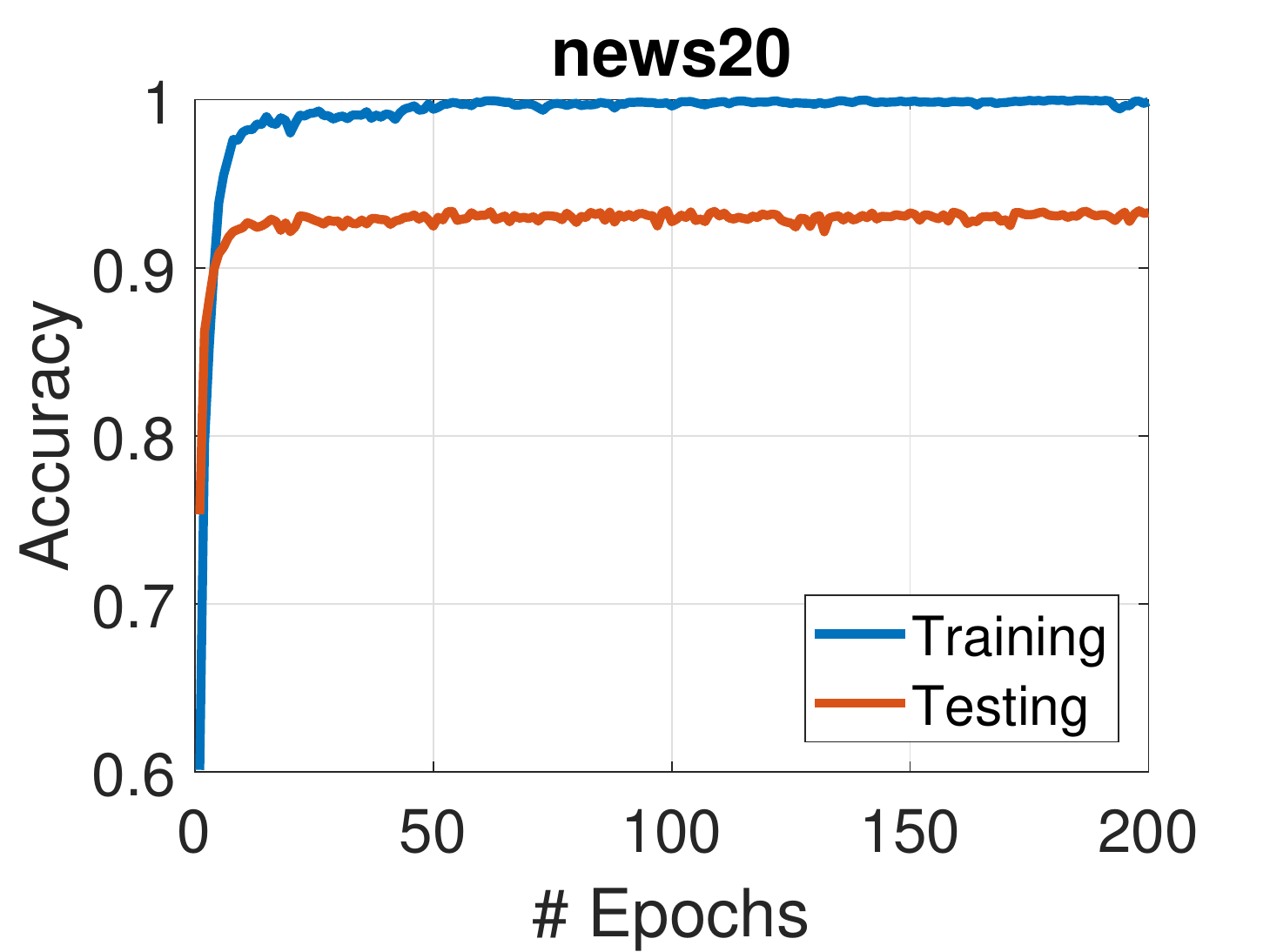}}
%			\centerline{\footnotesize{(b) Sampling using Eq. \ref{eqn:eta_t}}}
		\end{center}
	\end{minipage}	
    \begin{minipage}[b]{0.325\linewidth}
		\begin{center}
			\centerline{\includegraphics[width=\columnwidth]{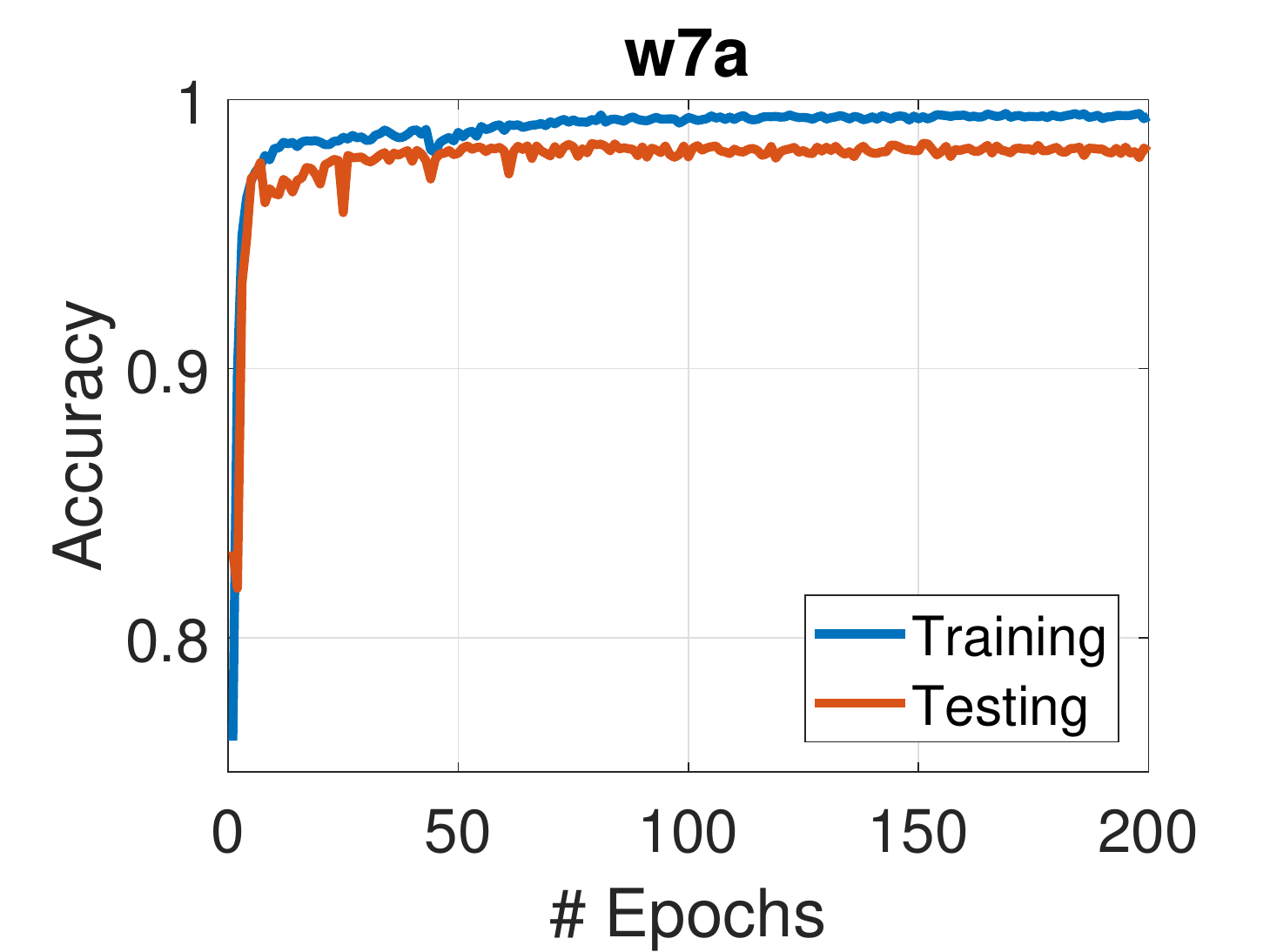}}
%			\centerline{\footnotesize{(b) Sampling using Eq. \ref{eqn:eta_t}}}
		\end{center}
	\end{minipage}	
    \vspace{-3mm}
	\caption{\footnotesize Illustration of training and testing {\bf (Top)} loss and {\bf (Bottom)} accuracy over epochs on different datasets.}\label{fig:binary-loss}
    %\vspace{-3mm}
\end{figure*}

\subsection{Binary Classification}

We first summarize our comparison results in Table \ref{tab:binary}. Overall our LMKL-Net outperforms all the other competitors, achieving the best in 3 out of 6 classes, while the other 3 best accuracies is obtained by LMKL. Compared with LMKL, however, LMKL-Net achieves {\bf 8.49\%} improvement on average in terms of accuracy. OBSCURE performs very closely to LMKL-Net.

%In fact all the three stochastic solvers perform very similarly. Note that both UNIFORM and MKL-Net utilize the average kernel for classification. The only difference is that MKL-Net learns the classifier parameters data-dependently while UNIFORM does not. This leads to {\bf 10.72\%} improvement 

In order to understand LMKL-Net, we illustrate the loss and accuracy on training and test data in Fig. \ref{fig:binary-loss}. We take for example adult-8, news20, and w7a. From the perspective of loss, we can observe clear overfitting in training on the first two datasets, as the testing loss increases with more epochs while the training loss decreases continuously. Surprisingly, we also observe that the quick and serious overfitting on adult-8 actually leads to very slow decrease in test accuracy, while on news20 such behavior is even hardly noticeable. This may explain why our solver performs worst on adult-8 among all the datasets. On the other hand, these observations also indicate the robustness of LMKL-Net in training and inference. In contrast, on w7a LMKL-Net is trained well, leading to the best performance among all the datasets. %Interestingly, by having more parameters in the network, GMKL-Net seems to work more robustly than MKL-Net with much less fluctuation

\begin{figure*}[t]	
    \begin{minipage}[b]{0.325\linewidth}
		\begin{center}
			\centerline{\includegraphics[width=.9\columnwidth]{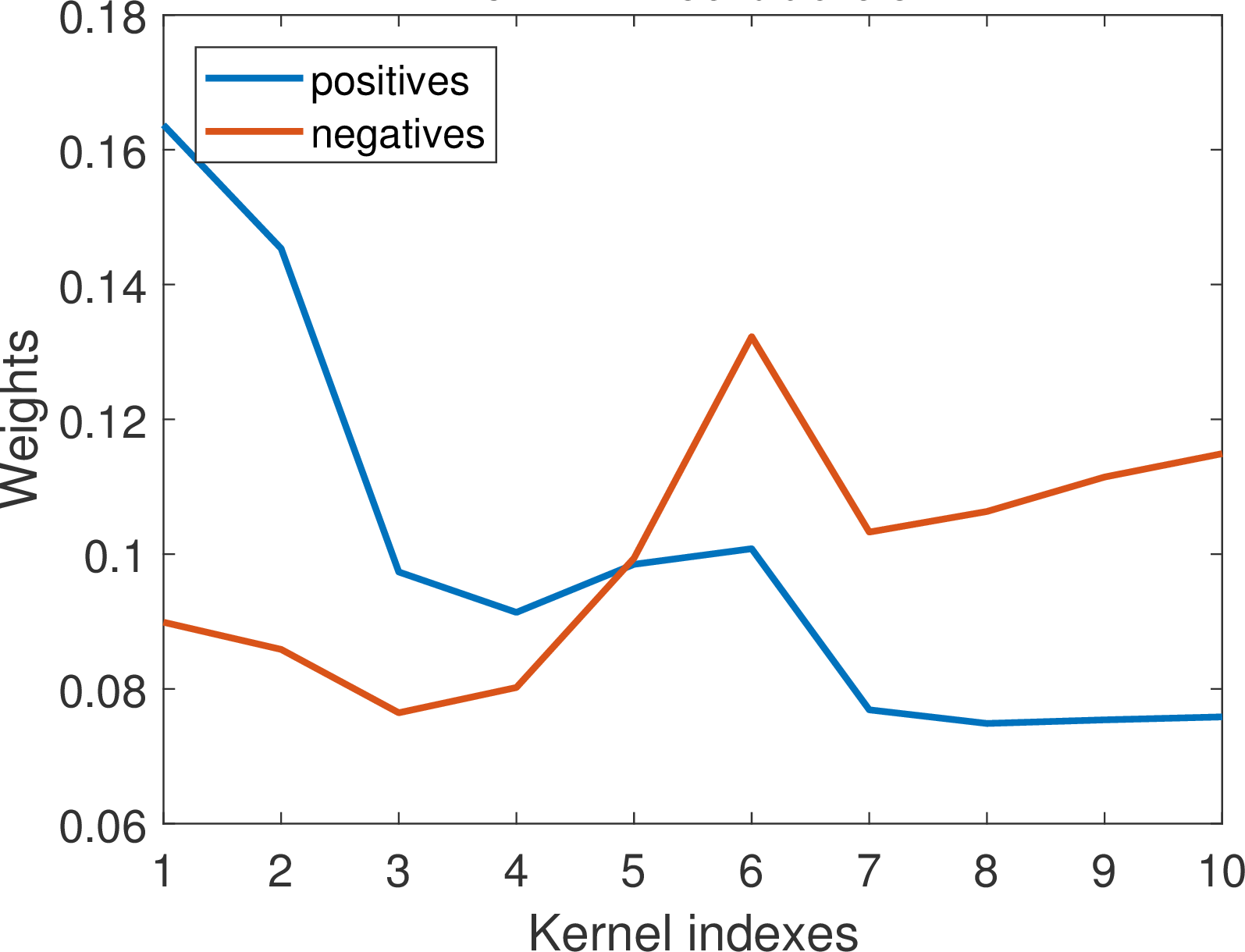}}
			\centerline{\footnotesize{(a) adult-8}}
		\end{center}
	\end{minipage}
	\begin{minipage}[b]{0.325\linewidth}
		\begin{center}
			\centerline{\includegraphics[width=.9\columnwidth]{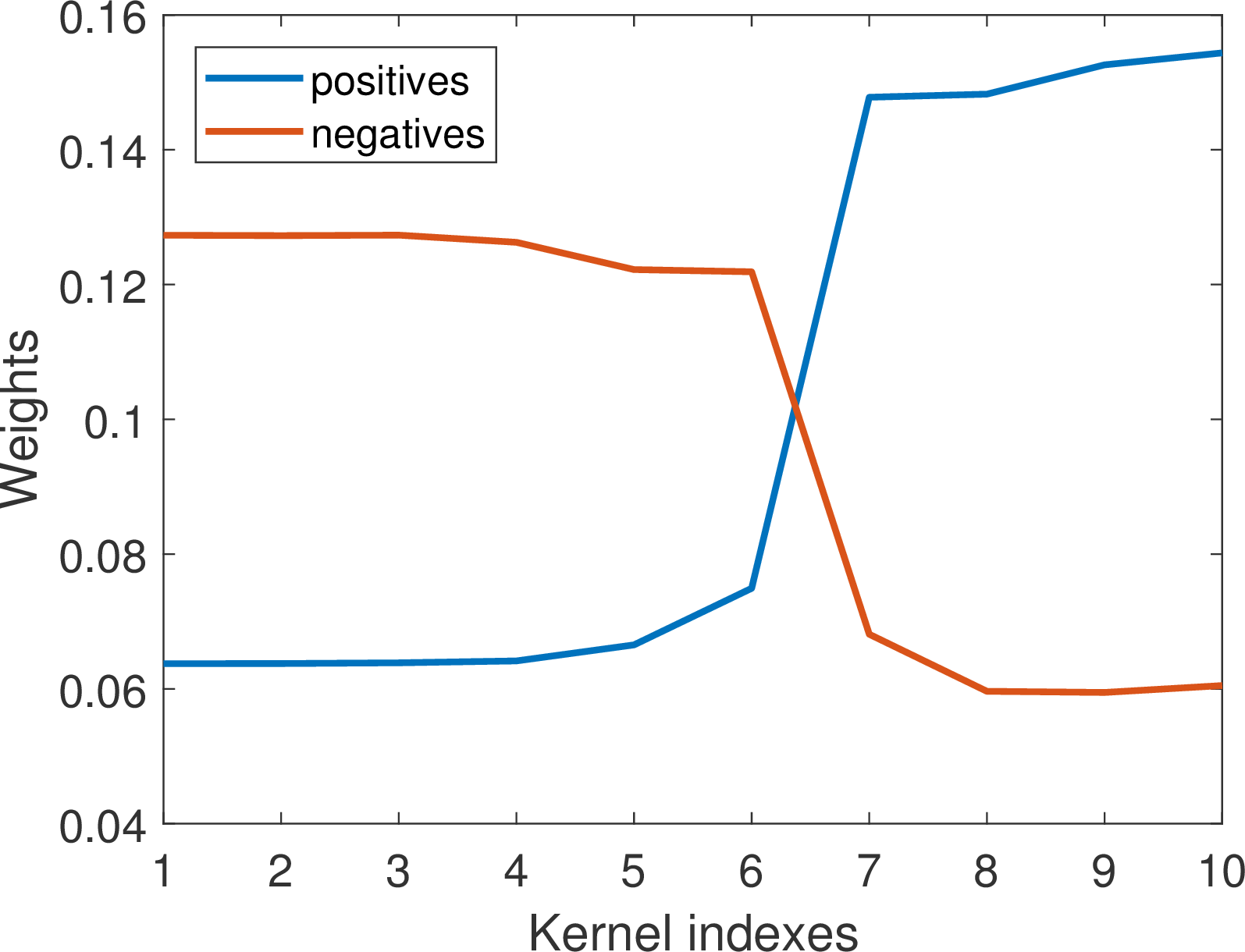}}
			\centerline{\footnotesize{(b) news20}}
		\end{center}
	\end{minipage}	
%     \begin{minipage}[b]{0.325\linewidth}
% 		\begin{center}
% 			\centerline{\includegraphics[width=\columnwidth]{weights/GMKL-Net-phishing.eps}}
% %			\centerline{\footnotesize{(a) Sampling using Eq. \ref{eqn:GD-sampler}}}
% 		\end{center}
% 	\end{minipage}
	\begin{minipage}[b]{0.325\linewidth}
		\begin{center}
			\centerline{\includegraphics[width=.9\columnwidth]{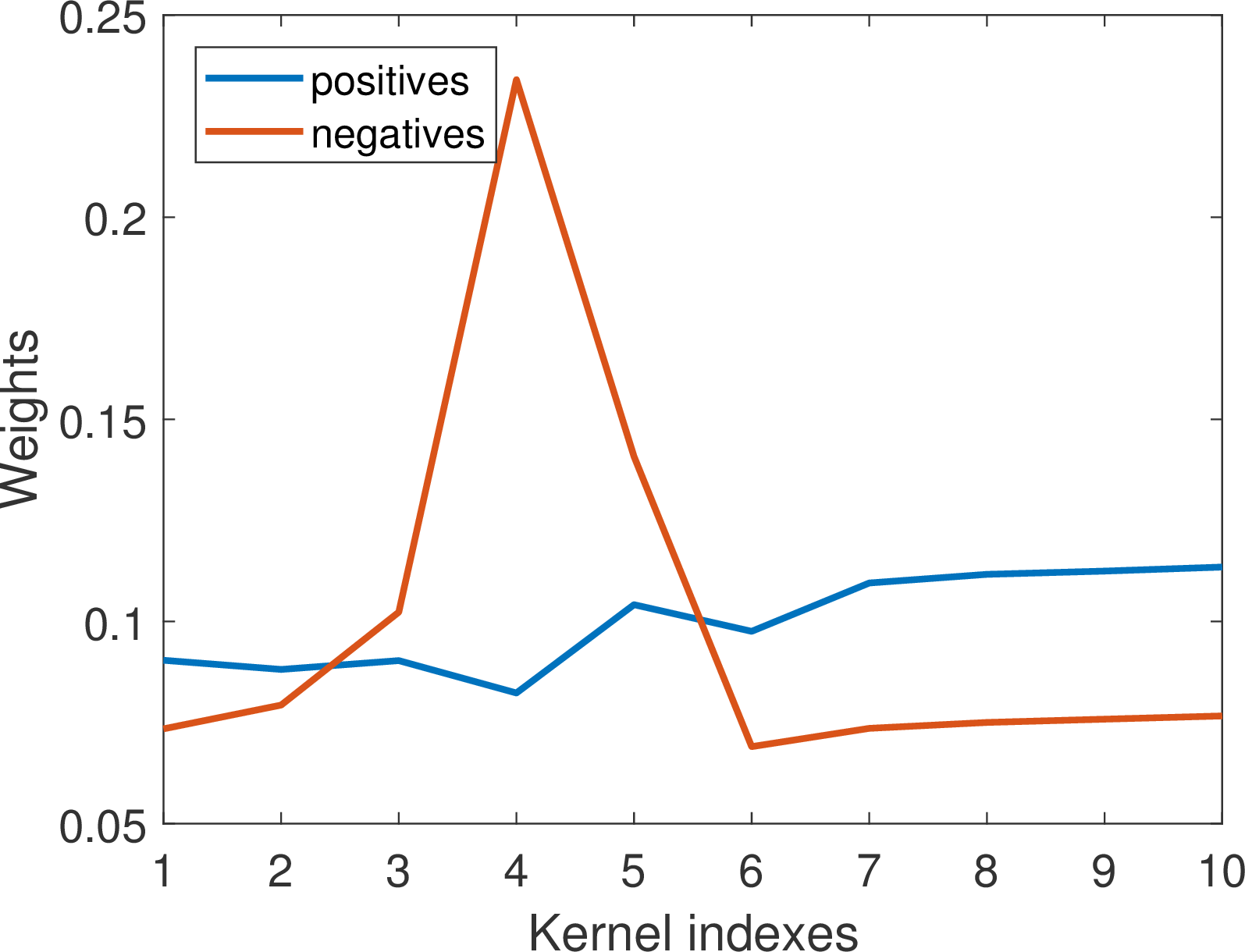}}
            \centerline{\footnotesize{(c) w7a}}
		\end{center}
	\end{minipage}	
    \vspace{-3mm}
	\caption{\footnotesize Illustration of learned mean kernel weights by averaging over the test data in each binary dataset. }\label{fig:binary-weights}
    \vspace{-4mm}
\end{figure*}

We illustrate the kernel weights learned by LMKL-Net in Fig. \ref{fig:binary-weights} as well. To do so, we first marginalize $h(\mathbf{K}(z); \omega)$ in Eq. \ref{eqn:h} for each sample $z$ over the training samples $x_i, \forall i$ to compute an $M$-dim vector, and then compute the mean vectors over positive and negative samples in each test dataset, respectively. As we expect from the nature of data dependency, the patterns on each dataset are quite distinct. For instance, on news20 positives prefer the last four kernels while negatives prefer the rest. The variance in kernel weights is relatively small compared with the mean. Therefore for clarification we only show the mean values.

\subsection{Multiclass Classification}
\begin{figure*}[t]
	\begin{minipage}[b]{0.325\linewidth}
		\begin{center}
			\centerline{\includegraphics[width=\columnwidth]{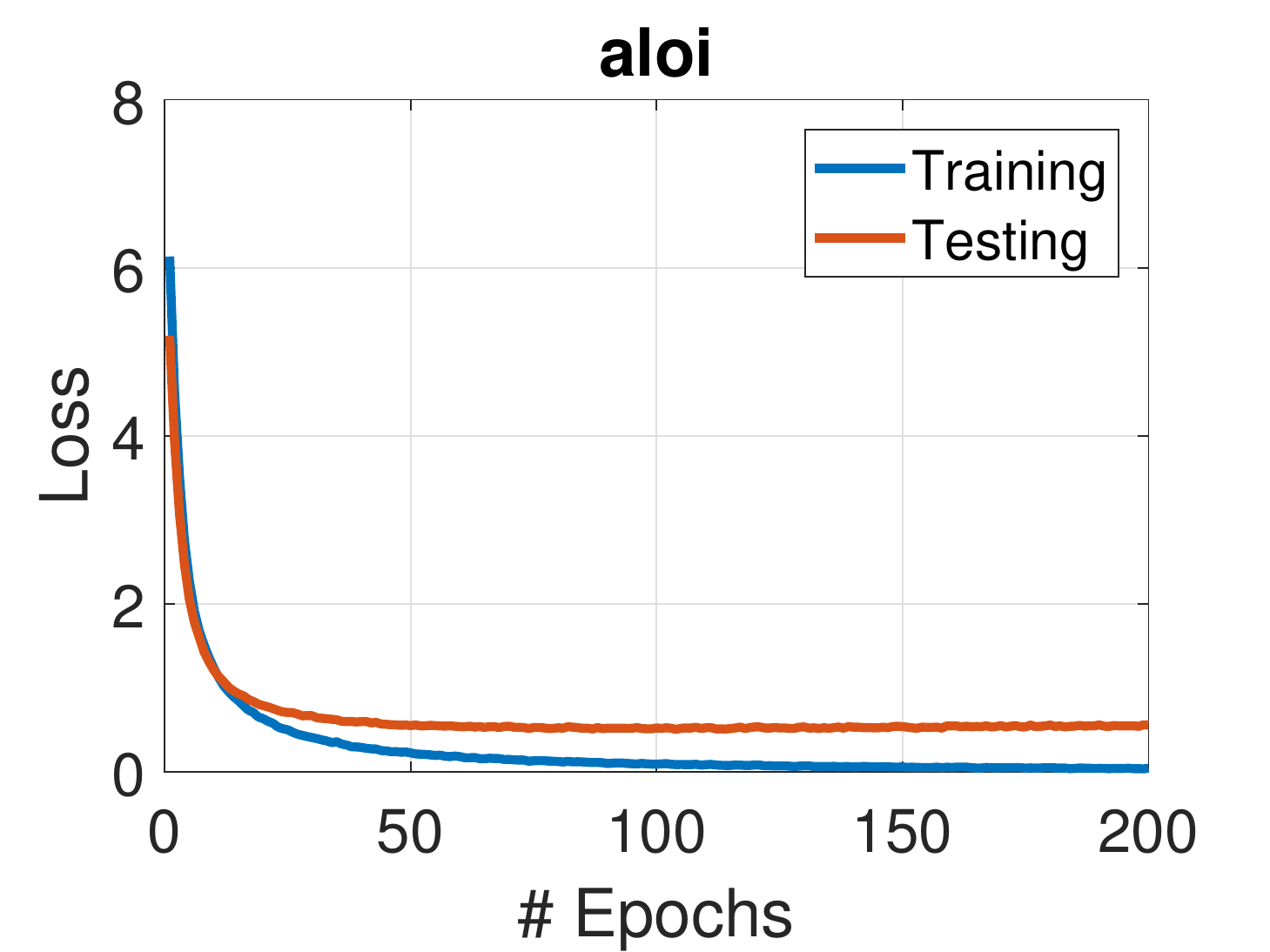}}
%			\centerline{\footnotesize{(b) Sampling using Eq. \ref{eqn:eta_t}}}
		\end{center}
	\end{minipage}
    \begin{minipage}[b]{0.325\linewidth}
		\begin{center}
			\centerline{\includegraphics[width=\columnwidth]{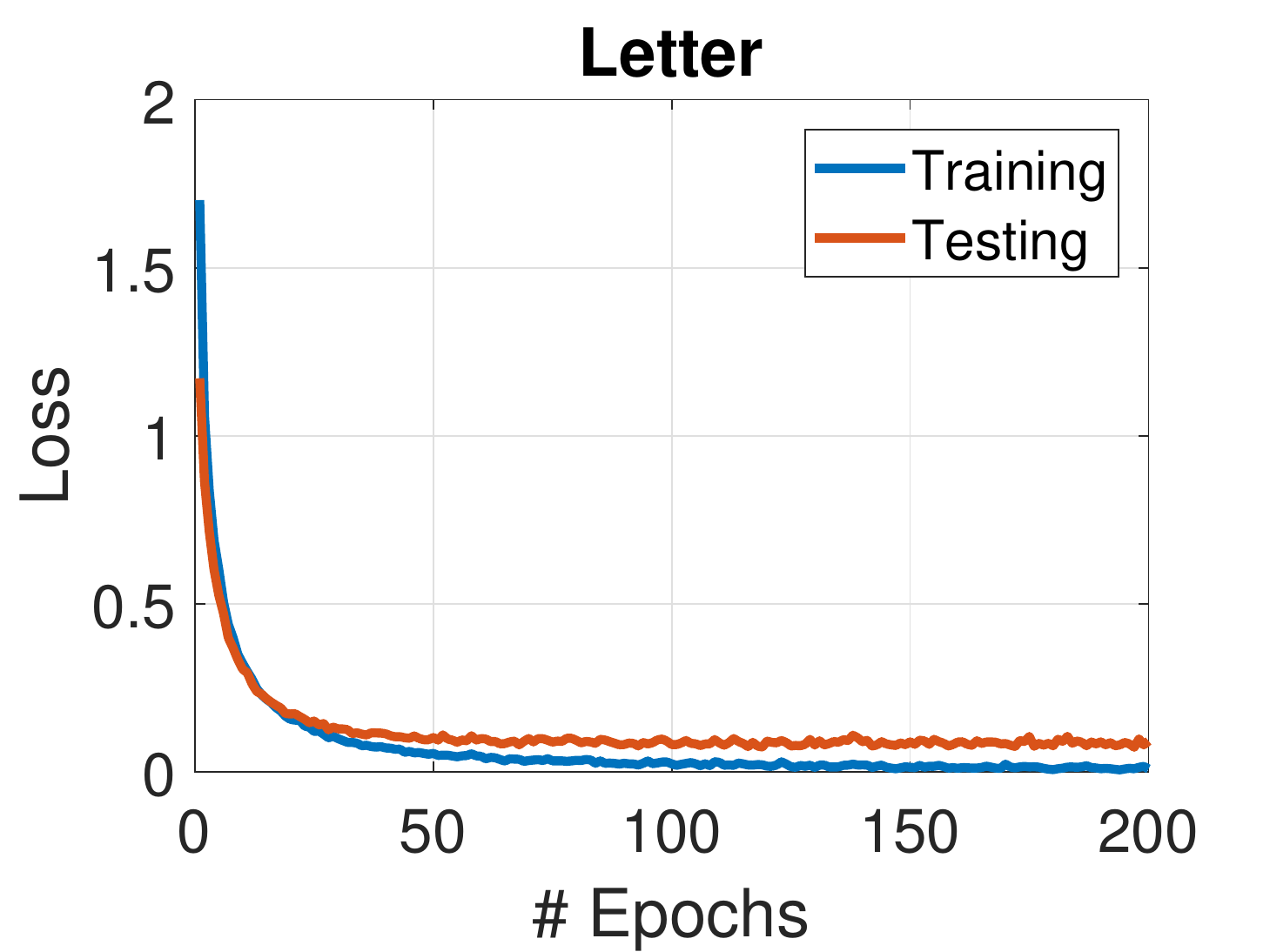}}
%			\centerline{\footnotesize{(a) Sampling using Eq. \ref{eqn:GD-sampler}}}
		\end{center}
	\end{minipage}
	\begin{minipage}[b]{0.325\linewidth}
		\begin{center}
			\centerline{\includegraphics[width=\columnwidth]{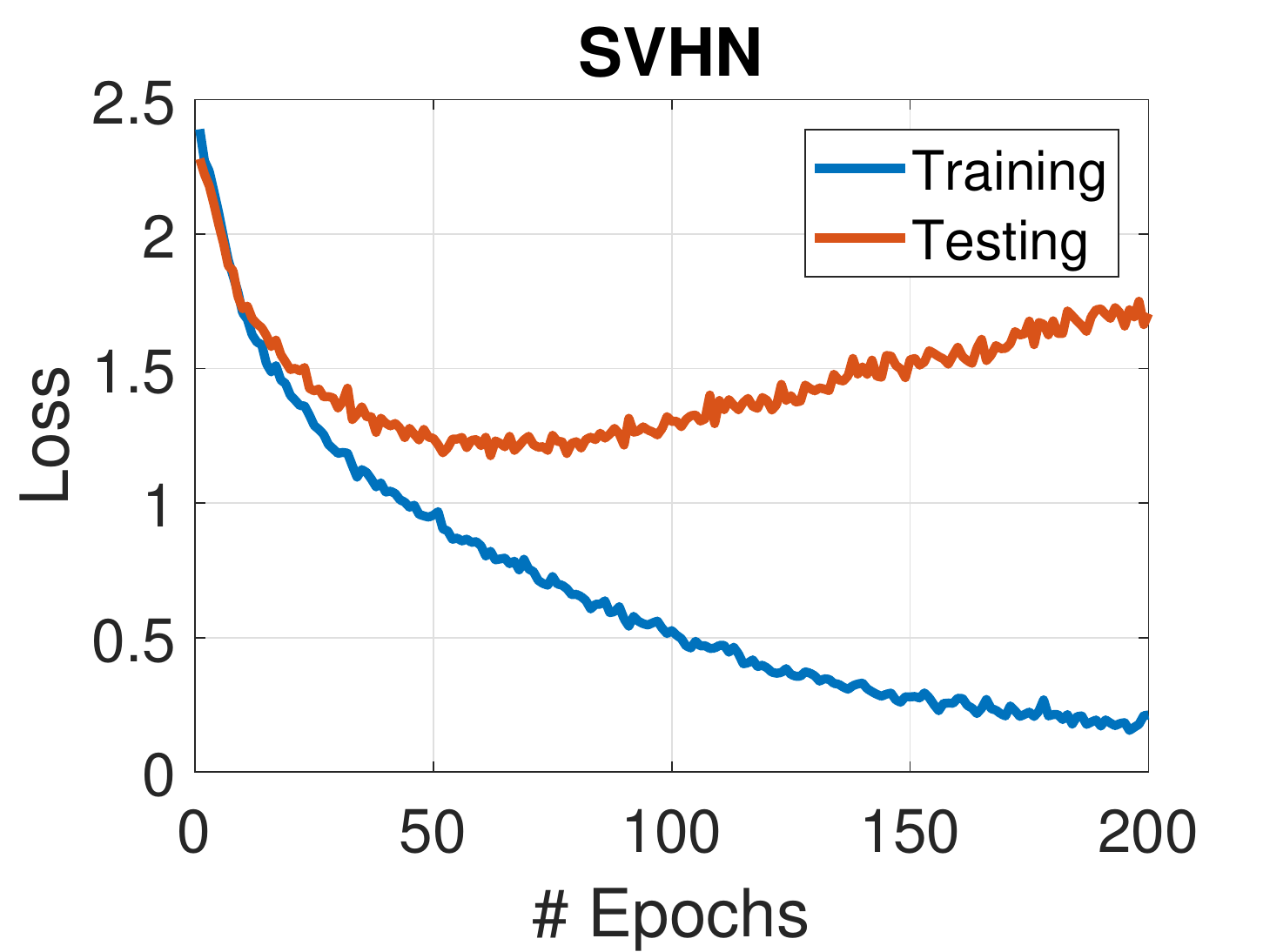}}
%			\centerline{\footnotesize{(b) Sampling using Eq. \ref{eqn:eta_t}}}
		\end{center}
	\end{minipage}	
    \begin{minipage}[b]{0.325\linewidth}
		\begin{center}
			\centerline{\includegraphics[width=\columnwidth]{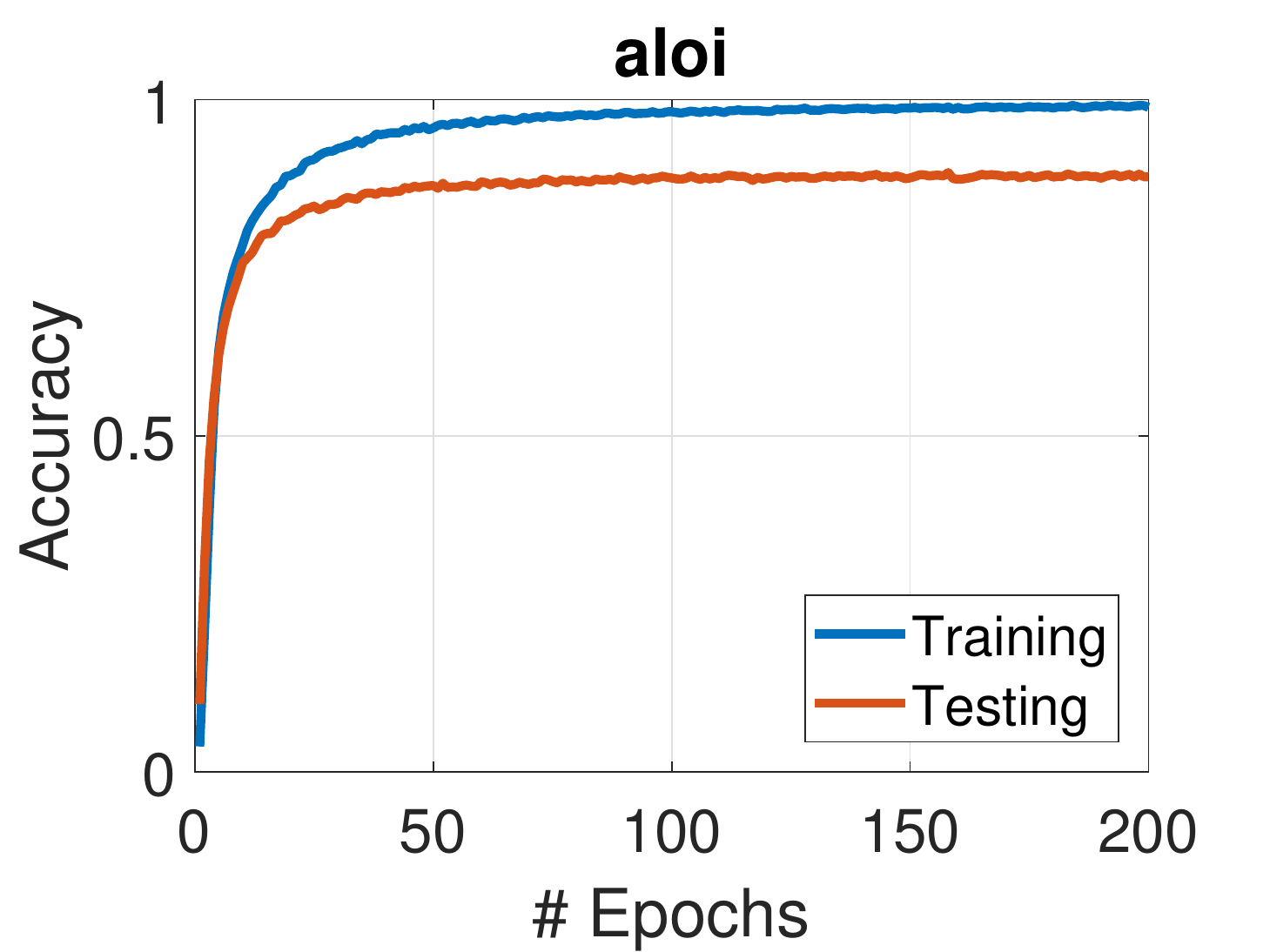}}
%			\centerline{\footnotesize{(b) Sampling using Eq. \ref{eqn:eta_t}}}
		\end{center}
	\end{minipage}
    \begin{minipage}[b]{0.325\linewidth}
		\begin{center}
			\centerline{\includegraphics[width=\columnwidth]{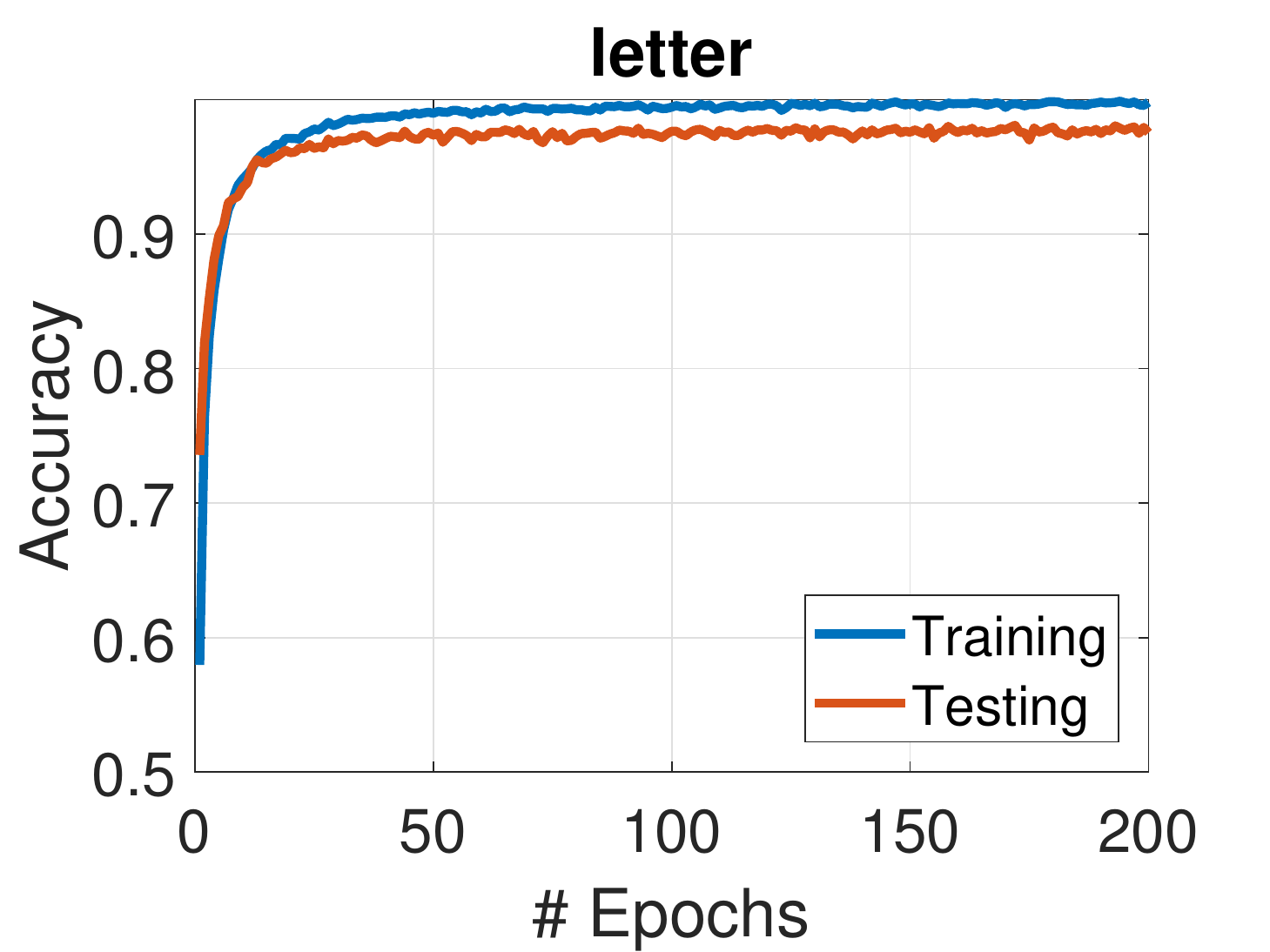}}
%			\centerline{\footnotesize{(a) Sampling using Eq. \ref{eqn:GD-sampler}}}
		\end{center}
	\end{minipage}
	\begin{minipage}[b]{0.325\linewidth}
		\begin{center}
			\centerline{\includegraphics[width=\columnwidth]{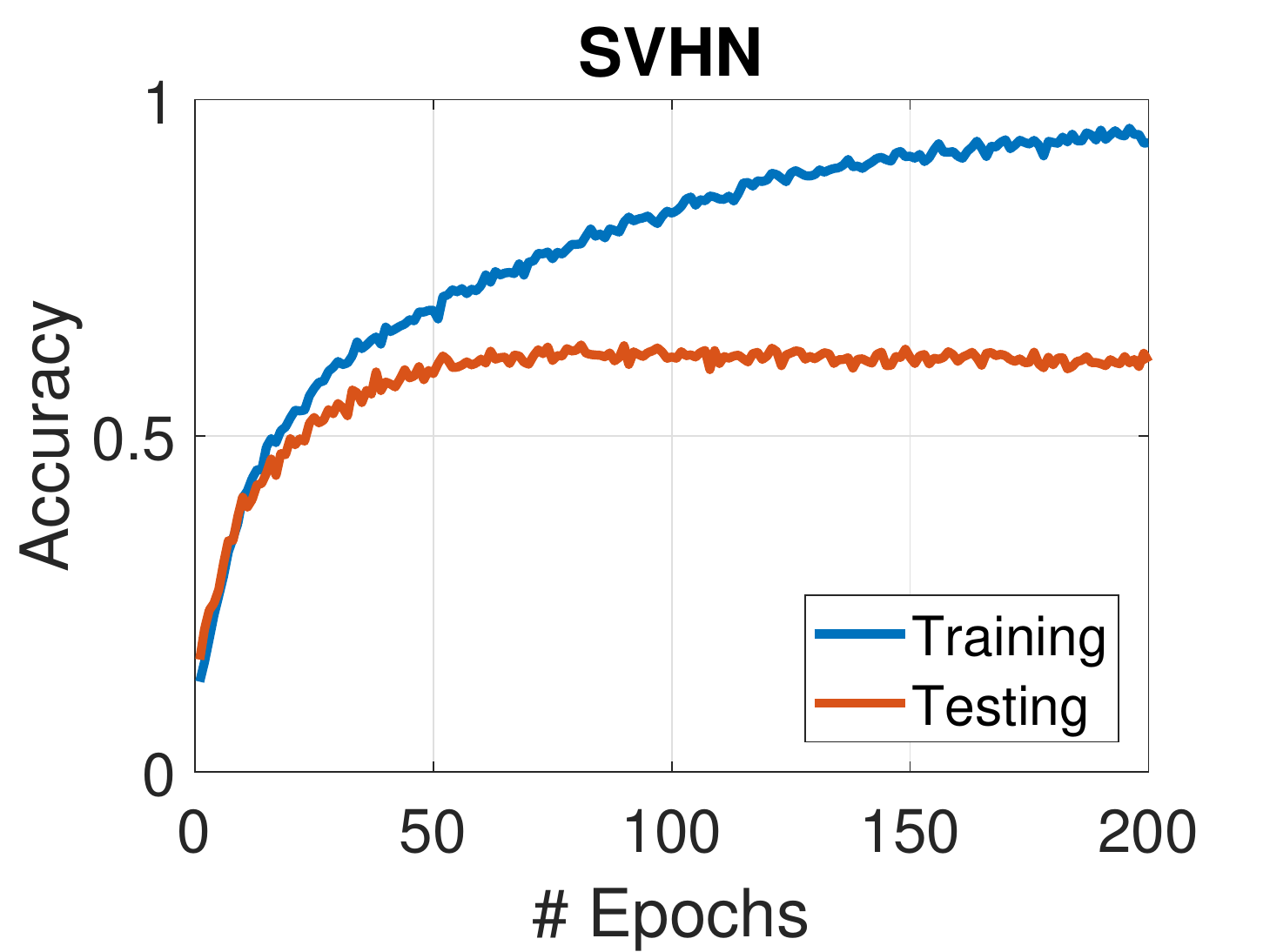}}
%			\centerline{\footnotesize{(b) Sampling using Eq. \ref{eqn:eta_t}}}
		\end{center}
	\end{minipage}	
%     \begin{minipage}[b]{\linewidth}
% 		\begin{center}
% 			\centerline{\includegraphics[width=\columnwidth]{weights/letter.eps}}
% %			\centerline{\footnotesize{(b) Sampling using Eq. \ref{eqn:eta_t}}}
% 		\end{center}
% 	\end{minipage}	
    \vspace{-3mm}
	\caption{\footnotesize Illustration of {\bf (top)} loss and {\bf (bottom)} accuracy in multiclass classification.}\label{fig:multiclass}
    \vspace{-3mm}
\end{figure*}

\begin{wraptable}{r}{7cm}\small
	\vspace{-15pt}
	\begin{center}    
		\begin{tabular}{|c|c|c|c|}
        	\hline & UNIFORM & OBSCURE & {\bf Ours} \\			
            \hline aloi & 86.74 & {\bf 89.60$\pm$0.14} & 89.37$\pm$0.16 \\
            \hline covtype & 82.12 & {\bf 83.56$\pm$0.07} & 82.48$\pm$0.09 \\
            \hline letter & 97.68 & 97.70$\pm$0.02 & {\bf 98.35$\pm$0.06} \\
            \hline protein & {\bf 70.24} & 69.95$\pm$0.11 & 69.92$\pm$0.27 \\
            \hline sensit & {\bf 85.20} & 83.76$\pm$0.05 & 84.93$\pm$0.15 \\
            \hline sensorless & 98.77 & 99.23$\pm$0.02 & {\bf 99.68$\pm$0.06} \\
            \hline shuttle & 99.19 & 99.84$\pm$0.01 & {\bf 99.87$\pm$0.01} \\
            \hline SVHN & 59.73 & 55.73$\pm$0.10 & {\bf 63.62$\pm$0.28} \\
            \hline\hline average & 84.96 & 84.92 & {\bf 86.03} \\ 
			\hline
		\end{tabular}
	\end{center}
%    \vspace{-3mm}
	\caption{\footnotesize Comparison on multiclass classification ($\%$).}
	\label{tab:multiclass}
    \vspace{-15pt}
\end{wraptable}
We summarize the comparison results in Table~\ref{tab:multiclass}. Again LMKL-Net performs the best overall with {\bf 1.11\%} improvement over OBSCURE.

We also illustrate the training and testing behavior of our solver in Fig. \ref{fig:multiclass}. The observations here on loss and accuracy are quite similar to those in binary classification. Moreover it seems that the gaps between the training and testing curves for both loss and accuracy are consistent with each other. Namely, it is very likely that smaller gaps in loss will result in better training and testing accuracy, and vice versa. 

We also test LMKL-Net on VGG MKL dataset as an example of using small-scale datasets. This dataset is created based on Caltech-101 where there are 101 object classes plus a background class. In our experiments we utilize the setup where for each class 15 images are sampled as training data and another 15 images as test data. 10 kernels are created as input. On this dataset our solver achieves $75.2\%$ with {\bf 0.6\%} improvement over the best number shown on the web page.

\subsection{Memory Footprint \& Computational Time}
Typically it needs about 20GB hard disk to store the precomputed kernels per dataset. Due to the batch size of 256, we need about $256*20/20000=0.26$GB memory to store the data for SGD training. Compared with the memory for loading all the precomputed kernels, the ratio will be, roughly speaking, $0.26/20={\bf 0.013}$. Empirically Since we observe that even we project the input kernels to a lower dimensional space than 256D, our solver can still achieve similar performance. This indicates that the active memory or ratio can be much smaller dependent on the network architectures.
 
%It is well known that network training using SGD has advantages over conventional methods for large-scale learning in terms of computational efficiency. 
For computational time we take aloi dataset consisting of 1K classes as an example for large-scale learning. Our test machine has a GTX 1080 graphics card (GPU) and an i7-6850K@3.6GHz CPU. It takes more than 10 hours to learn OBSCURE with multi-threads on CPU. Our LMKL-Net is trained on GPU, and both need about 6s to traverse the dataset once (\ie one epoch with about 20K samples). As we show in Fig. \ref{fig:binary-loss} and Fig.~\ref{fig:multiclass}, our solver usually can converge empirically within 50 epochs. Based on these statistics, we can roughly compute the running time ratio as $6*50/(3600*10)={\bf 0.0083}$. On the other datasets we do observe that the training time of OBSCURE seems scalable with the number of classes given similar sizes of training data. For instance OBSCURE needs less than 5 minutes to be trained for binary classification (but still the fastest among the other solvers). In contrast, LMKL-Net has similar training speed across different datasets as long as the sizes of training data are roughly the same. %It is worthy of mentioning that the running time of LMKL-Net depends on the network architectures and training setting. 

%Therefore, typically in large-scale learning our solvers can run about two orders of magnitude faster than other state-of-the-art MKL solvers with two orders of magnitude smaller memory, while achieving good classification accuracy.
 
\section{Conclusion}
In this paper we propose a deep neural network, LMKL-Net, as an efficient solver for localized multiple kernel learning (LMKL) problems. LMKL-Net consists of two major components, \ie an attentional network (AN) for learning the gating function and a multilayer perceptron (MLP) for learning a multiclass classifier in LMKL. We expect that the network can approximate the optimal functions in terms of accuracy given the input kernels. Empirically we demonstrate the performance of LMKL-Net on several benchmark datasets and compare it with some state-of-the-art MKL solvers. Overall LMKL-Net outperforms its competitors for both binary and multiclass classification. The robustness in training speed and the characteristic of SGD differentiate LMKL-Net from the other solvers, leading to about two orders of magnitude faster with much smaller memory footprint for large-scale learning.

%In this paper we propose a new learning problem in MKL, namely Bilinear Localized MKL (BLMKL). Accordingly we propose a simple feedforward neural network, MKL-Net, as an efficient solver to learn the bilinear localized prediction functions in BLMKL. We show that we can explicitly express the variables in BLMKL based on the network architectures, and thus the models learned from MKL-Net are indeed the solutions of BLMKL. We test MKL-Net as well as GMKL-Net, a generalized network, on both binary and multiclass classification. With much faster running speed and much smaller memory footprint, the overall performance of MKL-Net achieves the best among all the competitors.
	
\newpage
{\small
\bibliographystyle{ieee}
\bibliography{egbib}

\begin{thebibliography}{10}\itemsep=-1pt

\bibitem{alioschamultiple}
M.~Alioscha-Perez, M.~Oveneke, D.~Jiang, and H.~Sahli.
\newblock Multiple kernel learning via multi-epochs svrg.
\newblock In {\em 9th NIPS Workshop on Optimization for Machine Learning}, 12
  2016.

\bibitem{bach2009exploring}
F.~R. Bach.
\newblock Exploring large feature spaces with hierarchical multiple kernel
  learning.
\newblock In {\em NIPS}, pages 105--112, 2009.

\bibitem{bach2004multiple}
F.~R. Bach, G.~R. Lanckriet, and M.~I. Jordan.
\newblock Multiple kernel learning, conic duality, and the smo algorithm.
\newblock In {\em ICML}, page~6, 2004.

\bibitem{DBLP:journals/corr/abs-1709-10441}
B.~Bohn, M.~Griebel, and C.~Rieger.
\newblock A representer theorem for deep kernel learning.
\newblock {\em CoRR}, abs/1709.10441, 2017.

\bibitem{bottou2016optimization}
L.~Bottou, F.~E. Curtis, and J.~Nocedal.
\newblock Optimization methods for large-scale machine learning.
\newblock {\em arXiv preprint arXiv:1606.04838}, 2016.

\bibitem{bucak2014multiple}
S.~S. Bucak, R.~Jin, and A.~K. Jain.
\newblock Multiple kernel learning for visual object recognition: A review.
\newblock {\em TPAMI}, 36(7):1354--1369, 2014.

\bibitem{cortes2009learning}
C.~Cortes, M.~Mohri, and A.~Rostamizadeh.
\newblock Learning non-linear combinations of kernels.
\newblock In {\em NIPS}, pages 396--404, 2009.

\bibitem{pascal-voc-2009}
M.~Everingham, L.~Van~Gool, C.~K.~I. Williams, J.~Winn, and A.~Zisserman.
\newblock The {PASCAL} {V}isual {O}bject {C}lasses {C}hallenge 2009 {(VOC2009)}
  {R}esults.
\newblock
  http://www.pascal-network.org/challenges/VOC/voc2009/workshop/index.html.

\bibitem{gehler2009feature}
P.~Gehler and S.~Nowozin.
\newblock On feature combination for multiclass object classification.
\newblock In {\em ICCV}, pages 221--228, 2009.

\bibitem{gonen08icml}
M.~G\"{o}nen and E.~Alpayd{\i}n.
\newblock Localized multiple kernel learning.
\newblock In {\em ICML}, 2008.

\bibitem{gonen2011multiple}
M.~G{\"o}nen and E.~Alpayd{\i}n.
\newblock Multiple kernel learning algorithms.
\newblock {\em JMLR}, 12(Jul):2211--2268, 2011.

\bibitem{gonen2011multitask}
M.~G{\"o}nen, M.~Kandemir, and S.~Kaski.
\newblock Multitask learning using regularized multiple kernel learning.
\newblock In {\em Neural Information Processing}, pages 500--509, 2011.

\bibitem{han2012probability}
Y.~Han and G.~Liu.
\newblock Probability-confidence-kernel-based localized multiple kernel
  learning with $ l\_ $\{$p$\}$ $ norm.
\newblock {\em IEEE Transactions on Systems, Man, and Cybernetics, Part B
  (Cybernetics)}, 42(3):827--837, 2012.

\bibitem{jagarlapudi2009algorithmics}
S.~N. Jagarlapudi, G.~Dinesh, S.~Raman, C.~Bhattacharyya, A.~Ben-Tal, and
  R.~Kr.
\newblock On the algorithmics and applications of a mixed-norm based kernel
  learning formulation.
\newblock In {\em NIPS}, pages 844--852, 2009.

\bibitem{Jain12}
A.~Jain, S.~V.~N. Vishwanathan, and M.~Varma.
\newblock Spg-gmkl: Generalized multiple kernel learning with a million
  kernels.
\newblock In {\em SIGKDD}, August 2012.

\bibitem{jawanpuria2011multi}
P.~Jawanpuria and J.~S. Nath.
\newblock Multi-task multiple kernel learning.
\newblock In {\em ICDM}, pages 828--838, 2011.

\bibitem{jawanpuria2015generalized}
P.~Jawanpuria, J.~S. Nath, and G.~Ramakrishnan.
\newblock Generalized hierarchical kernel learning.
\newblock {\em JMLR}, 16(1):617--652, 2015.

\bibitem{ji2009multi}
S.~Ji, L.~Sun, R.~Jin, and J.~Ye.
\newblock Multi-label multiple kernel learning.
\newblock In {\em NIPS}, pages 777--784, 2009.

\bibitem{kingma2014adam}
D.~P. Kingma and J.~Ba.
\newblock Adam: A method for stochastic optimization.
\newblock {\em arXiv preprint arXiv:1412.6980}, 2014.

\bibitem{kloft2011lp}
M.~Kloft, U.~Brefeld, S.~Sonnenburg, and A.~Zien.
\newblock Lp-norm multiple kernel learning.
\newblock {\em JMLR}, 12(Mar):953--997, 2011.

\bibitem{krogh1992simple}
A.~Krogh and J.~A. Hertz.
\newblock A simple weight decay can improve generalization.
\newblock In {\em NIPS}, pages 950--957, 1992.

\bibitem{lei2016localized}
Y.~Lei, A.~Binder, U.~Dogan, and M.~Kloft.
\newblock Localized multiple kernel learning—a convex approach.
\newblock In {\em ACML}, pages 81--96, 2016.

\bibitem{li2017triply}
X.~Li, B.~Gu, S.~Ao, H.~Wang, and C.~X. Ling.
\newblock Triply stochastic gradients on multiple kernel learning.
\newblock In {\em UAI}, 2017.

\bibitem{liu2014sample}
X.~Liu, L.~Wang, J.~Zhang, and J.~Yin.
\newblock Sample-adaptive multiple kernel learning.
\newblock In {\em AAAI}, pages 1975--1981, 2014.

\bibitem{long2017pde}
Z.~Long, Y.~Lu, X.~Ma, and B.~Dong.
\newblock Pde-net: Learning pdes from data.
\newblock {\em arXiv preprint arXiv:1710.09668}, 2017.

\bibitem{martins2011online}
A.~F.~T. Martins, N.~Smith, E.~Xing, P.~Aguiar, and M.~Figueiredo.
\newblock Online learning of structured predictors with multiple kernels.
\newblock In {\em AISTATS}, pages 507--515, 2011.

\bibitem{meirom2016nuc}
E.~Meirom and P.~Kisilev.
\newblock Nuc-mkl: A convex approach to non linear multiple kernel learning.
\newblock In {\em AISTATS}, pages 610--619, 2016.

\bibitem{mhaskar2016deep}
H.~N. Mhaskar and T.~Poggio.
\newblock Deep vs. shallow networks: An approximation theory perspective.
\newblock {\em Analysis and Applications}, 14(06):829--848, 2016.

\bibitem{moeller2014geometric}
J.~Moeller, P.~Raman, S.~Venkatasubramanian, and A.~Saha.
\newblock A geometric algorithm for scalable multiple kernel learning.
\newblock In {\em AIStats}, pages 633--642, 2014.

\bibitem{moeller2016unified}
J.~Moeller, S.~Swaminathan, and S.~Venkatasubramanian.
\newblock A unified view of localized kernel learning.
\newblock In {\em ICDM}, pages 252--260, 2016.

\bibitem{mu2011non}
Y.~Mu and B.~Zhou.
\newblock Non-uniform multiple kernel learning with cluster-based gating
  functions.
\newblock {\em Neurocomputing}, 74(7):1095--1101, 2011.

\bibitem{murugesan2017multi}
K.~Murugesan and J.~Carbonell.
\newblock Multi-task multiple kernel relationship learning.
\newblock In {\em ICDM}, pages 687--695, 2017.

\bibitem{nair2010rectified}
V.~Nair and G.~E. Hinton.
\newblock Rectified linear units improve restricted boltzmann machines.
\newblock In {\em ICML}, pages 807--814, 2010.

\bibitem{ijcai2017-758}
K.~Nguyen.
\newblock Nonparametric online machine learning with kernels.
\newblock In {\em IJCAI}, pages 5197--5198, 2017.

\bibitem{orabona2010online}
F.~Orabona, L.~Jie, and B.~Caputo.
\newblock Online-batch strongly convex multi kernel learning.
\newblock In {\em CVPR}, pages 787--794, 2010.

\bibitem{orabona2012multi}
F.~Orabona, L.~Jie, and B.~Caputo.
\newblock Multi kernel learning with online-batch optimization.
\newblock {\em JMLR}, 13(Feb):227--253, 2012.

\bibitem{orabona2011ultra}
F.~Orabona and J.~Luo.
\newblock Ultra-fast optimization algorithm for sparse multi kernel learning.
\newblock In {\em ICML}, 2011.

\bibitem{UNIFORM}
P.~Pavlidis, J.~Weston, J.~Cai, and W.~N. Grundy.
\newblock Gene functional classification from heterogeneous data.
\newblock In {\em Proceedings of the fifth annual international conference on
  Computational biology}, pages 249--255, 2001.

\bibitem{rakotomamonjy2008simplemkl}
A.~Rakotomamonjy, F.~R. Bach, S.~Canu, and Y.~Grandvalet.
\newblock Simplemkl.
\newblock {\em JMLR}, 9(Nov):2491--2521, 2008.

\bibitem{sirignano2017dgm}
J.~Sirignano and K.~Spiliopoulos.
\newblock Dgm: A deep learning algorithm for solving partial differential
  equations.
\newblock {\em arXiv preprint arXiv:1708.07469}, 2017.

\bibitem{song2017optimizing}
H.~Song, J.~J. Thiagarajan, P.~Sattigeri, and A.~Spanias.
\newblock Optimizing kernel machines using deep learning.
\newblock {\em arXiv preprint arXiv:1711.05374}, 2017.

\bibitem{sonnenburg2006large}
S.~Sonnenburg, G.~R{\"a}tsch, C.~Sch{\"a}fer, and B.~Sch{\"o}lkopf.
\newblock Large scale multiple kernel learning.
\newblock {\em JMLR}, 7(Jul):1531--1565, 2006.

\bibitem{strobl2013deep}
E.~V. Strobl and S.~Visweswaran.
\newblock Deep multiple kernel learning.
\newblock In {\em ICMLA}, volume~1, pages 414--417, 2013.

\bibitem{tang2009multiple}
L.~Tang, J.~Chen, and J.~Ye.
\newblock On multiple kernel learning with multiple labels.
\newblock In {\em IJCAI}, pages 1255--1260, 2009.

\bibitem{varma2009more}
M.~Varma and B.~R. Babu.
\newblock More generality in efficient multiple kernel learning.
\newblock In {\em ICML}, pages 1065--1072, 2009.

\bibitem{vedaldi2009multiple}
A.~Vedaldi, V.~Gulshan, M.~Varma, and A.~Zisserman.
\newblock Multiple kernels for object detection.
\newblock In {\em ICCV}, pages 606--613, 2009.

\bibitem{Vishy10}
S.~V.~N. Vishwanathan, Z.~Sun, N.~Theera-Ampornpunt, and M.~Varma.
\newblock Multiple kernel learning and the {SMO} algorithm.
\newblock In {\em NIPS}, December 2010.

\bibitem{weinan2017deep}
E.~Weinan, J.~Han, and A.~Jentzen.
\newblock Deep learning-based numerical methods for high-dimensional parabolic
  partial differential equations and backward stochastic differential
  equations.
\newblock {\em Communications in Mathematics and Statistics}, 5(4):349--380,
  2017.

\bibitem{yang2009group}
J.~Yang, Y.~Li, Y.~Tian, L.~Duan, and W.~Gao.
\newblock Group-sensitive multiple kernel learning for object categorization.
\newblock In {\em ICCV}, pages 436--443, 2009.

\bibitem{zien2007multiclass}
A.~Zien and C.~S. Ong.
\newblock Multiclass multiple kernel learning.
\newblock In {\em ICML}, pages 1191--1198, 2007.

\end{thebibliography}
}  
	
\end{document}